# Random Drift Particle Swarm Optimization

**Jun Sun** SUNJUN_WX@HOTMAIL.COM
Key Laboratory of Advanced Process Control for Light Industry (Ministry of Education), Jiangnan University, No 1800, Lihu Avenue, Wuxi, Jiangsu 214122

**Xiaojun Wu** WU_XIAOJUN@YAHOO.COM.CN
Key Laboratory of Advanced Process Control for Light Industry (Ministry of Education), Jiangnan University, No 1800, Lihu Avenue, Wuxi, Jiangsu 214122

**Vasile Palade** VASILE.PALADE@CS.OX.AC.UK
Department of Computer Science, University of Oxford, Parks Road, Oxford, OX1 3QD, United Kingdom

**Wei Fang** WXFANGWEI@HOTMAIL.COM
Key Laboratory of Advanced Process Control for Light Industry (Ministry of Education), Jiangnan University, No 1800, Lihu Avenue, Wuxi, Jiangsu 214122

**Yuhui Shi** YUHUI.SHI@XJTLU.EDU.CN
Department of Computer Science, University of Oxford, Parks Road, Oxford, OX1 3QD, United Kingdom

***Abstract***

The random drift particle swarm optimization (RDPSO) algorithm, inspired by the free electron model in metal conductors placed in an external electric field, is presented, systematically analyzed and empirically studied in this paper. The free electron model considers that electrons have both a thermal and a drift motion in a conductor that is placed in an external electric field. The motivation of the RDPSO algorithm is described first, and the velocity equation of the particle is designed by simulating the thermal motion as well as the drift motion of the electrons, both of which lead the electrons to a location with minimum potential energy in the external electric field. Then, a comprehensive analysis of the algorithm is made, in order to provide a deep insight into how the RDPSO algorithm works. It involves a theoretical analysis and the simulation of the stochastic dynamical behavior of a single particle in the RDPSO algorithm. The search behavior of the algorithm itself is also investigated in detail, by analyzing the interaction between the particles. Some variants of the RDPSO algorithm are proposed by incorporating different random velocity components with different neighborhood topologies. Finally, empirical studies on the RDPSO algorithm are performed by using a set of benchmark functions from the CEC2005 benchmark suite. Based on the theoretical analysis of the particle's behavior, two methods of controlling the algorithmic parameters are employed, followed by an experimental analysis on how to select the parameter values, in order to obtain a good overall performance of the RDPSO algorithm and its variants in real-world applications. A further performance comparison between the RDPSO algorithms and other variants of PSO is made to prove the efficiency of the RDPSO algorithms.

## 1. Introduction

Particle swarm optimization (PSO) is a population-based optimization method attributed to be originally developed by Kennedy and Eberhart (Kennedy and Eberhart, 1995; Eberhart and Kennedy, 1995). It is widely known that PSO is rooted in two paradigms (Kennedy and Eberhart, 1995). One obvious root is its ties with artificial life in general, and bird flocking, fish schooling, and swarm theory in particular. The other root is associated with evolutionary algorithms (EAs), such as genetic algorithms (GAs) and evolutionary programming (EP). However, unlike EAs, PSO has no evolution operators similar to crossover and selection. PSO optimizes a problem by iteratively improving a population of candidate solutions with respect to an objective (fitness) function. The candidate solutions, called particles, move through the problem space according to simple mathematical formulae describing the particles' position and velocity. The movement of each particle is influenced by its own experiences, and is also guided towards the current best known position.

During the last decade, PSO has gained increasing popularity due to its effectiveness in performing difficult optimization tasks. The reason why PSO is attractive is that it gets better solutions, in a faster and cheaper way compared to other methods, whereas has fewer parameters to adjust. It has been successfully used in many research and application areas. An extensive survey of PSO applications can be found in (Poli, 2007; 2008).

To gain insights into how the algorithm works, some researchers have theoretically analyzed the PSO algorithm. These analyses mainly aimed for the behavior of the individual particle in the PSO algorithm, which is essential to the understanding of the search mechanism of the algorithm and to the parameter selection (Kennedy, 1998; Ozcan, and Mohan, 1999; Clerc and Kennedy, 2002; van den Bergh, 2002; Eberhart and Shi, 1998; Trelea, 2003; Emara, and Fattah, 2004; Gavi and Passino, 2003; Kadirkamanathan, et al, 2006; Jiang, 2007; Solis and Wets, 1981). For example, Kennedy analysed a simplified particle behavior and demonstrated different particle trajectories for a range of design choices (Kennedy, 1998). Clerc and Kennedy undertook the first formal analysis of the particle trajectory and its stability properties (Clerc and Kennedy, 2002). As for the algorithm itself, Van den Bergh proved that the canonical PSO is not a global search algorithm (van den Bergh, 2002), even not a local one, by using the convergence criterion provided by Solis and Wets (Solis and Wets, 1981).

In addition to the analyses mentioned above, there has been a considerable amount of work performed in improving the original version of the PSO through empirical studies. The original PSO proposed in (Kennedy and Eberhart, 1995) appeared to have weak local search ability, due to the slow convergence speed of the particles. It is universally known that the tradeoff between the local search (exploitation) and the global search (exploration) is vital for the performance of the algorithm. As such, the original PSO needs to accelerate the convergence speed of the particles in order to achieve a better balance between exploitation and exploration. The work in this area, first carried out by Shi and Eberhart, involves introducing an inertia weight into the update equation for velocities (Shi and Eberhart, 1998). Clerc proposed another acceleration method by adding a constriction factor in the velocity update equation, in order to release the restriction on the particle's velocity during the convergence history (Clerc, 1999). The acceleration techniques were shown to work well, and the above two variants of PSO have laid the foundation for further enhancement of the PSO algorithm.

In the original PSO, the PSO with inertia weight (PSO-In) and the PSO with constriction factor (PSO-Co), the search of the particles is guided by the global best position and their personal best positions. In these versions of PSO, all particles are neighbors of each other so that their neighborhood topology is known as the global best topology or the global best model. Although the algorithm with this model is able to efficiently obtain the best approximate solutions for many problems, it is more prone to encounter premature convergence when solving harder problems. If the global best particle sticks to a local or suboptimal point, it would mislead the other particles to move towards that point. In other words, other promising search areas might be missed. This had led to the investigation of other neighborhood topologies known as the local best models, first studied by Eberhart and Kennedy (1995) and subsequently in depth by many other researchers (Suganthan, 1999; Kennedy, 1999; 2002; Liang and Suganthan, 2005; Mendes, et al., 2004; Parrott and Li, 2006; Bratton and Kennedy, 2007; Kennedy and Mendes, 2002; van den Bergh and Engelbrecht, 2004; Lane et al., 2004; Li, 2004). The objective there was to find other possible topologies to improve the performance of the PSO algorithm.

In PSO, the particle essentially follows a semi-deterministic trajectory defined by a velocity update formula with two random acceleration coefficients. This is a semi-deterministic search, which restricts the search domain of each particle and may weaken the global search ability of the algorithm, particularly at the later stage of the search process. In view of this limitation, some researchers have proposed several probabilistic PSO algorithms, which simulate the particle trajectories by direct sampling, using a random number generator, or from a distribution of practical interests (Kennedy, 2003; 2004; Sun, et al., 2012; Krohling, 2004; Secrest and Lamon, 2003; Richer and Blackwell, 2006; Kennedy, 2006). The Bare Bones PSO (BBPSO) family is a typical class of probabilistic PSO algorithms (Kennedy, 2003). In BBPSO, each particle does not have a velocity vector, but its new position is sampled "around" a supposedly good one, according to a certain probability distribution, such as the Gaussian distribution in the original version (Kennedy, 2003). Several other new BBPSO variants used other distributions which seem to generate better results (Kennedy, 2004; 2006).

Researchers also turned to hybrid algorithms that incorporate other search methods into the PSO algorithm, for the purpose of playing to the advantages of different optimization algorithms (Angeline, 1998; Løvbjerg et al., 2001; Zhang and Xie, 2003; Devicharan and Mohan, 2004; Chen et al., 2007; Settles and Soule, 2005; Higashi and Iba, 2003; Pant et al., 2008). Angeline undertook the first work in this area by introducing a tournament selection into the PSO, based on the particle's current fitness, so that the properties that make some solutions superior were transferred directly to some of the less effective particles (Angeline, 1998). Besides, some researchers introduced various efficient strategies into the PSO in order to enhance the search ability of the algorithm (Riget and Vesterstroem, 2002; Lovbjerg and Krink, 2002; Xie et al., 2002; Krink et al., 2002; Liang et al., 2006). For instance, Liang et al. proposed a PSO with a novel learning strategy, in which all other particles' historical best information is used to update a particle's velocity. It was shown that this strategy can diversify the swarm to avoid premature convergence (Liang et al., 2006).

In this paper, based on a random drift model, we present a new version of PSO, which is called the random drift particle swarm optimization (RDPSO). This PSO variant is inspired by the free electron model in metal conductors in an external electric field. The model considers that each electron in a conductor, which is situated in an external electric field, has both a thermal motion as well as a drift motion (Omar, 1993). The drift motion is caused by the electric field and is the directional movement of the electron in the opposite direction to the electric field. On the other hand, the thermal motion is random in essence, and it exists even in the absence of an external electric field. The two motions together bring the electron into a location with minimum potential energy, which is analogous to the process of searching for the optimal solution to an optimization problem. Our motivation of designing the RDPSO algorithm, based on this model, was to improve the search ability of the PSO algorithm by only modifying the update equation of the particle's velocity, instead of by revising the algorithm based on the update equation of the canonical PSO, which would probably increase the complexity of the algorithm and its computational cost.

The basics of the original concept of the random drift model for PSO were sketched in our previous work (Sun et al., 2010). In the initial limited version of the algorithm, the velocity of the particle's drift motion is simply expressed by the summation of the cognition part and the social part in the velocity update equation of the original PSO, which is not consistent with the physical meaning of the random drift model. This paper is to propose a more concise form for the drift velocity, which is more in line with the physical meaning of the model, and a novel strategy for determining the random velocity, and thus a new version of the RDPSO algorithm. In order to gain an in-depth understanding of how the RDPSO works, we make comprehensive theoretical analyses of the behavior of the individual particle in the RDPSO and the search behavior of the algorithm. Four variants of the RDPSO are proposed based on different random velocities and neighborhood topologies. Comprehensive empirical studies on the RDPSO algorithm by using the CEC2005 benchmark suite are performed to verify the effectiveness of the algorithm.

To this end, the paper firstly describes the principle of the RDPSO and analyzes the behavior of the single particle in the RDPSO. The motivations of the RDPSO from a trajectory analysis point of view together with the free electron model are formulated in detail, and the random drift model for the RDPSO is presented. Based on this model, the velocity of the particle is assumed to be the superimposition of the random velocity component and the drift velocity component, which reflect the global search as well as the local search of the particle, respectively. The mathematical expressions of the two velocity components and, subsequently, the update equation for the particle's velocity are given. After that, the conditions for the particle's position to be probabilistically bounded are theoretically derived, and are later verified by stochastic simulations on the particle's behavior.

Then, the search mechanism of the RDPSO algorithm is investigated. The effects of the thermal and drift motions on the particle's search behavior are analyzed. The drift velocity component leads the particle to move toward its personal best position and the global best position as well, and thus it essentially implements the local search of the particle. The random velocity component makes the particle more volatile and its position is pulled or pushed away from the global best position, reflecting the global search of the particle. The interactions between the particles in the swarm are also analyzed in order to show that the RDPSO algorithm may provide a good balance between the global and the local search. Next, four RDPSO variants are proposed based on the combination of the two topologies (i.e., the global best model, and the ring neighborhood topology for the local best model) with two strategies for the random velocity components (i.e., one that uses the mean best position to determine the random velocity component, and the other one that employs randomly selected personal best position to compute the random velocity component).

Finally, empirical studies on the RDPSO algorithm are undertaken by using the CEC2005 benchmark suite. The issues of the parameter control and selection with respect to the thermal and drift coefficients of the algorithm are addressed by testing the algorithm with different parameter settings on three benchmarks. Then, the parameter settings that are identified to result in good algorithmic performance are further tested and compared on the first twelve functions of the CEC2005 benchmark suite. For each RDPSO variant, the parameter settings that yield good overall algorithmic performance are found out. The RDPSO variants with the identified parameter configurations and some other PSO variants are tested by all the twenty five problems of the benchmark suite in order to make a thorough performance comparison and verify the efficiency of the RDPSO algorithms.

The remainder of the paper is organized as follows. Section 2 gives a brief introduction to the PSO algorithm. Section 3 presents the motivation, the procedure and the analyses of the RDPSO algorithm as well as its variants. Empirical studies on the parameter selection for the RDPSO algorithm and the performance comparison are provided in Section 4. Finally, the paper is concluded in Section 5.

## 2. Particle Swarm Optimization

In a PSO with $M$ individuals, each individual is treated as a volume-less particle in the $N$-dimensional space, with the current position vector and the velocity vector of particle $i$ at the $n^{\text{th}}$ iteration represented as $X_{i,n}=(X_{i,n}^1,X_{i,n}^2,\cdots,X_{i,n}^N)$ and $V_{i,n}=(V_{i,n}^1,V_{i,n}^2,\cdots,V_{i,n}^N)$, respectively. The particle moves according to the following equations:

$$V_{i,n+1}^j = V_{i,n}^j + c_1 r_{i,n}^j (P_{i,n}^j - X_{i,n}^j) + c_2 R_{i,n}^j (G_n^j - X_{i,n}^j), \tag{1}$$

$$X_{i,n+1}^j = X_{i,n}^j + V_{i,n+1}^j, \tag{2}$$

for $i=1,2,\cdots M; j=1,2\cdots,N$, where $c_1$ and $c_2$ are known as the acceleration coefficients. The vector $P_{i,n}=(P_{i,n}^1,P_{i,n}^2,\cdots,P_{i,n}^N)$ is the best previous position (the position giving the best objective function value or fitness value) of particle *i,* called the personal best (*pbest*) position, and the vector $G_n = (G_n^1, G_n^2, \cdots, G_n^N)$ is the position of the best particle among all the particles in the population and called the global best (*gbest*) position. Without loss of generality, we consider the following minimization problem:

$$\text{Minimize} \quad f(X), s.t. \quad X \in S \subseteq R^N, \tag{3}$$

where $f(X)$ is an objective function and $S$ is the feasible space. Accordingly, $P_{i,n}$ can be updated by

$$P_{i,n} = \begin{cases} X_{i,n} & \text{if} \quad f(X_{i,n}) < f(P_{i,n-1}) \\ P_{i,n-1} & \text{if} \quad f(X_{i,n}) \geq f(P_{i,n-1}) \end{cases}, \tag{4}$$

and $G_n$ can be found by $G_n = P_{g,n}$, where $g = \arg\min_{1 \leq i \leq M}[f(P_{i,n})]$. The parameters $r_{i,n}^j$ and $R_{i,n}^j$ are sequences of two different random numbers distributed uniformly on (0, 1), which is denoted by $r_{i,n}^j, R_{i,n}^j \sim U(0,1)$. Generally, the value of $V_{i,n}^j$ is restricted within the interval $[-V_{\max}, V_{\max}]$.

The original PSO algorithm with equation (1) appears to have a weak local search ability. It should be noted that the tradeoff between the local search (exploitation) and the global search (exploration) is vital for the performance of the algorithm. Therefore, the original PSO needs to accelerate the convergence speed of the particles in order to achieve a better balance between exploitation and exploration. Work in this area, first carried out by Shi and Eberhart [16], involves introducing an inertia weight into equation (1), and the resulting update equation for velocities becomes:

$$V_{i,n+1}^j = wV_{i,n}^j + c_1 r_{i,n}^j (P_{i,n}^j - X_{i,n}^j) + c_2 R_{i,n}^j (G_n^j - X_{i,n}^j), \tag{5}$$

where *w* is the inertia weight. The PSO algorithm with equation (5) replacing equation (1) is known as the PSO with inertia weight (PSO-In). The inertia weight *w* can be a positive value chosen according to experience or from a linear or nonlinear function of the iteration number. When *w* is 1, the PSO-In is equivalent to the original PSO. The values of $c_1$ and $c_2$ in equation (5) are generally set to be 2 as originally recommended by Kennedy and Eberhart (1995), which implies that the 'social' and 'cognition' parts have the same influence on the velocity update.

Clerc (1999) proposed another acceleration method by adding a constriction factor in the velocity update equation (1) in order to ensure the convergence of the PSO without imposing any restriction on velocities, as given below.

$$V_{i,n+1}^j = \chi[V_{i,n}^j + c_1 r_{i,n}^j (P_{i,n}^j - X_{i,n}^j) + c_2 R_{i,n}^j (G_n^j - X_{i,n}^j)], \tag{6}$$

where the constant $\chi$ is known as the constriction factor and is determined by

$$\chi = \frac{2}{\left|2-\varphi-\sqrt{\varphi^2-4\varphi}\right|}, \quad \varphi = c_1 + c_2 . \tag{7}$$

This version of PSO is known as the PSO with constriction factor (PSO-Co). It was shown by Clerc and Kennedy (Clerc and Kennedy, 2002) that the swarm shows stable convergence if $\varphi \geq 4$. If $\chi \in [0,1]$, the approach is very similar to the concept of the inertia weight with $w = \chi$, $c_1' = \chi c_1$, $c_2' = \chi c_2$. Clerc and Kennedy recommended a value of 4.1 for the sum of $c_1$ and $c_2$, which leads to $\chi = 0.7298$ and $c_1$=$c_2$=2.05 (Clerc and Kennedy, 2002).

These two versions of the PSO algorithm collectively referred to as the canonical PSO algorithms accelerate the convergence speed of the swarm effectively and have better performance than the original PSO in general. They have laid the foundation for further enhancement of the PSO. There are many other versions of the PSO algorithm as have been mentioned in Section 1, but most of them are based on these two versions.

# 3. Random Drift Particle Swarm Optimization (RDPSO)

## 3.1 The Motivation and Procedure of RDPSO

In (Clerc and Kennedy, 2002), the trajectory analysis demonstrated that the convergence of the whole particle swarm may be achieved if each particle converges to its local focus, $p_{i,n} = (p_{i,n}^1, p_{i,n}^2, \cdots p_{i,n}^N)$ defined at the coordinates

$$p_{i,n}^j = \frac{c_1 r_{i,n}^j P_{i,n}^j + c_2 R_{i,n}^j G_n^j}{c_1 r_{i,n}^j + c_2 R_{i,n}^j}, \quad 1 \leq j \leq N . \tag{8}$$

In fact, as the particles are converging to their own local attractors, their current positions, *pbest* positions, local focuses and the *gbest* position are all converging to one point. This way, the canonical PSO algorithm is said to be convergent. Since $p_{i,n}$ is a random point uniformly distributed within the hyper-rectangle with $P_{i,n}$ and $G_n$ being the two ends of its diagonal, the particle's directional movement towards $p_{i,n}$ makes the particle search around this hyper-rectangle and improves its fitness value locally. Hence, this directional movement essentially reflects the local search of the particle. In equation (1), (5) or (6), there are three parts on the right side. The last two ones are known as the 'cognition' part and the 'social' part, the superimposition of which results in the directional motion of the particle toward $p_{i,n}$. The first part on the right side of each equation is the 'inertia part', which may lead the particle to fly away from $p_{i,n}$ or $G_n$ and provide necessary momentum for the particle to search globally in the search space. The 'inertia part' is deterministic and reflects the global search of the particle.

The motivation of the proposed RDPSO algorithm comes from the above trajectory analysis of the canonical PSO and the free electron model in metal conductors placed in an external electric field (Omar, 1993). According to this model, the movement of an electron is the superimposition of the thermal motion, which appears to be a random movement, and the drift motion (i.e., the directional motion) caused by the electric field. That is, the velocity of the electron can be expressed by $V = VR + VD$, where $VR$ and $VD$ are called the random velocity and the drift velocity, respectively. The random motion (i.e., the thermal motion) exists even in the absence of the external electric field, while the drift motion is a directional movement in the opposite direction of the external electric field. The overall physical effect of the electron's movement is that the electron careens towards the location of the minimum potential energy. In a non-convex-shaped metal conductor in an external electric field, there may be many locations of local minimum potential energies, which the drift motion generated by the electric force may drive the electron to. If the electron only had the drift motion, it might stick into a point of local minimum potential energy, just as a local optimization method converges to a local minimum of an optimization problem. The thermal motion can make the electron more volatile and, consequently, helps the electron to escape the trap of local minimum potential energy, just as a certain random search strategy is introduced into the local search technique to lead the algorithm to search globally. Therefore, the movement of the electron is a process of minimizing its potential energy. The goal of this process is essentially to find out the minimum solution of the minimization problem, with the position of the electron represented as a candidate solution and the potential energy function as the objective function of the problem.

Inspired by the above facts, we assume that the particle in the RDPSO behaves like an electron moving in a metal conductor in an external electric field. The movement of the particle is thus the superposition of the thermal and the drift motions, which implement the global search and the local search of the particle, respectively. The trajectory analysis, as described in the first paragraph of this subsection, indicates that, in the canonical PSO, the particle's directional movement toward its local attractor $p_{i,n}$ reflects the local search of the particle. In the proposed RDPSO, the drift motion of the particle is also defined as the directional movement toward $p_{i,n}$, which is the main inheritance of the RDPSO from the canonical PSO. However, in the RDPSO, the 'inertia part' in the velocity equation of the canonical PSO is replaced by the random velocity component. This is the main difference between the RDPSO and the canonical PSO. The thermal motion in the RDPSO is far different from the 'inertia' movement in the canonical PSO, although both of them have an identical functionality, namely, to implement the global search ability of the particle. From a physical perspective, the particle in the canonical PSO is in a mechanical movement and its velocity and acceleration can be depicted by a set of deterministic dynamics equations. Consequently, it is natural that there should be an 'inertia part' in the velocity equation to reflect the change in the particle's velocity with time. On the contrary, the thermal motion of the particle in the RDPSO is random in nature and can not be described by the dynamics equations used for mechanical movements. The only way of describing the thermal motion in statistical physics is by providing the probability distribution function of the particle's velocity or momentum. It is unnecessary and impossible to depict the exact change of the particle's velocity with time. Therefore, there is no 'inertia part' in the velocity equation of the RDPSO algorithm anymore. From an algorithm design point of view, the role of the 'inertia part' in the global search is assumed by the random velocity component in the RDPSO, so that there is no need of an 'inertia part' in the RDPSO. Therefore, the velocity of the particle in the RDPSO algorithm has two components, i.e., the thermal component and the drift component. Mathematically, the velocity of particle $i$ in the $j$th dimension can be expressed by $V_{i,n+1}^j = VR_{i,n+1}^j + VD_{i,n+1}^j$ ( $1 \le i \le M$ , $1 \le j \le N$ ), where $VR_{i,n+1}^j$ and $VD_{i,n+1}^j$ are the random velocity component and the drift velocity component, respectively.

A further assumption is that the value of the random velocity component $VR_{i,n+1}^j$ follows the Maxwell velocity distribution law. Consequently, $VR_{i,n+1}^j$ essentially follows a normal distribution (i.e., Gaussian distribution) whose probability density function is given by

$$f_{VR_{i,n+1}^j}(v) = \frac{1}{\sqrt{2\pi}\sigma_{i,n+1}^j} e^{\frac{-v^2}{2(\sigma_{i,n+1}^j)^2}}, \tag{9}$$

where $\sigma_{i,n+1}^j$ is the standard deviation of the distribution. Using stochastic simulation, we can express $VR_{i,n+1}^j$ as

$$VR_{i,n+1}^j = \sigma_{i,n+1}^j \varphi_{i,n+1}^j, \tag{10}$$

where $\varphi_{i,n+1}^j$ is a random number with a standard normal distribution, i.e., $\varphi_{i,n+1}^j \sim N(0,1)$. $\sigma_{i,n+1}^j$ must be determined in order to calculate $VR_{i,n+1}^j$. An adaptive strategy is adopted for $\sigma_{i,n+1}^j$:

$$\sigma_{i,n+1}^j = \alpha \,|\, C_n^j - X_{i,n}^j \,|, \tag{11}$$

where $C_n = (C_n^1, C_n^2, \cdots, C_n^N)$ is known as the mean best (*mbest*) position defined by the mean of the *pbest* positions of all the particles, namely,

$$C_n^j = (1/M)\sum_{i=1}^{M} P_{i,n}^j, \quad (1 \le j \le N). \tag{12}$$

Thus, equation (10) can be restated as

$$VR_{i,n+1}^j = \alpha \,|\, C_n^j - X_{i,n}^j \,|\, \varphi_{i,n+1}^j, \tag{13}$$

where $\alpha > 0$ is an algorithmic parameter called the thermal coefficient.

As for the drift velocity component, $VD_{i,n+1}^j$, its role is to achieve the local search of the particle. As has been mentioned above, the directional movement toward $p_{i,n}$ essentially plays this role. The original expression of $VD_{i,n+1}^j$ in (Sun et al., 2010) is just the combination of the 'cognitive part' and the 'social part' of equation (1), namely,

$$VD_{i,n+1}^j = c_1 r_{i,n}^j (P_{i,n}^j - X_{i,n}^j) + c_2 R_{i,n}^j (G_n^j - X_{i,n}^j),$$

and its more compact form is $VD_{i,n+1}^j = (c_1 r_{i,n}^j + c_2 R_{i,n}^j)(p_{i,n}^j - X_{i,n}^j)$, where $p_{i,n}^j$ is given by equation (8). Since the scaling factor $(c_1 r_{i,n}^j + c_2 R_{i,n}^j)$ is a random number, its effect is to make the movement of the particle randomized, which is not consistent with the free electron model. Therefore, in this paper we modify it to be the following simple linear expression:

$$VD_{i,n+1}^j = \beta (p_{i,n}^j - X_{i,n}^j), \tag{14}$$

where $\beta > 0$ is a deterministic constant and is another algorithmic parameter called the drift coefficient. Equation (14) has a clear physical meaning that it reflects the particle's directional movement towards $p_{i,n}$. In Theorem A1 in the Appendix, it is proven that, if there is only drift motion and, i.e., $V_{i,n+1}^j = VD_{i,n+1}^j$, $X_{i,n}^j \to p_{i,n}^j$ as $n \to \infty$ when $0 < \beta < 2$, meaning that the expression of $VD_{i,n+1}^j$ in equation (14) can indeed guarantee the particle's directional movement toward $p_{i,n}$ as an overall result. More specifically, if $0 < \beta < 1$, $X_{i,n}^j$ asymptotically converges to $p_{i,n}^j$, which means that the sampling space of $X_{i,n+1}$ does not cover the hyper-rectangle with $P_{i,n}$ and $G_n$ being the two ends of its diagonal. If $\beta = 1$, $X_{i,n+1}^j$ is identical to $p_{i,n}^j$ so that the sampling space of $X_{i,n+1}$ is exactly the hyper-rectangle. If $1 < \beta < 2$, $X_{i,n}^j$ converges to $p_{i,n}^j$ in oscillation and thus the sampling space of $X_{i,n+1}$ covers the hyper-rectangle and even other neighborhoods of $G_n$, where points with better fitness values may exist. As such, when we select the value of $\beta$ for real application of the RDPSO algorithm, it may be desirable to set $1 \le \beta < 2$ for good local search ability of the particles.

With the above specification, a novel set of update equations can be obtained for the particle of the RDPSO algorithm:

$$V_{i,n+1}^{j} = \alpha \,|\, C_n^j - X_{i,n}^j \,|\, \varphi_{i,n+1}^j + \beta (p_{i,n}^j - X_{i,n}^j), \tag{15}$$

$$X_{i,n+1}^{j} = X_{i,n}^{j} + V_{i,n+1}^{j}. \tag{16}$$

The procedure of the algorithm is outlined below. Like in the canonical PSO, the value of $V_{i,n}^j$ in the RDPSO is also restricted within the interval $[-V_{\max}, V_{\max}]$ at each iteration.

**Procedure of the RDPSO algorithm:**

**Begin**

Initialize the current positions and velocities of all the particles randomly;

Set the personal best position of each particle to be its current position;

Set *n*=0;

**While** (termination condition = false)

**Do**

*n*=*n*+1;

Compute mean best position $C_n$;

**for** (*i*=1 to *M*)

Evaluate the objective function value $f(X_{i,n})$;

Update $P_{i,n}$ according to equation (4) and then update $G_n$;

**for** *j*=1 to *N*

$V_{i,n+1}^{j} = \alpha \,|\, C_n^j - X_{i,n}^j \,|\, \varphi_{i,n}^j + \beta (p_{i,n}^j - X_{i,n}^j)$;

$X_{i,n+1}^{j} = X_{i,n}^{j} + V_{i,n+1}^{j}$;

**end for**

**end for**

**end do**

**end**

### 3.2. Dynamical Behavior of the RDPSO Particle

An analysis of the behavior of an individual particle in the RDPSO is very essential to understanding how the RDPSO algorithm works and how to select the algorithmic parameters. Since the particle's velocity is the superimposition of the thermal velocity and the drift velocity, the conditions for the particle's position to converge or to be bounded are far more complex than those given in subsection 3.1 when only the drift motion exists. In this subsection, we undertake theoretical and empirical studies on the stochastic dynamical behavior of the particle in the RDPSO. Since each dimension of the particle's position is updated independently, we only need to consider a single particle in the one-dimensional space without loss of generality. As such, equations (15) and (16) can be simplified as

$$V_{n+1} = \alpha\,|C - X_n|\,\varphi_{n+1} + \beta(p - X_n)\,, \tag{17}$$

$$X_{n+1} = X_{n+1} + V_{n+1}\,, \tag{18}$$

where $X_n$ and $V_n$ denote the current position and the velocity of the particle, respectively, and the local focus of the particle and the mean best position are denoted by $p$ and $C$, which are treated as probabilistically bounded random variables, i.e., $P\{\sup|p|<\infty\}=1$ and $P\{\sup|C|<\infty\}=1$. In equation (17), $\{\varphi_n\}$ is a sequence of independent identically distributed random variables with $\varphi_n \sim N(0,1)$.

Since the distribution of $\varphi_n$ is symmetrical with respect to the ordinate, equation (17) has the following equivalence:

$$V_{n+1} = \alpha(X_n - C)\varphi_{n+1} - \beta(X_n - p)\,, \tag{19}$$

that is, the probability distributions of $V_{n+1}$ in equations (17) and (19) are the same. Based on equations (19) and (18), several theorems on the dynamical behavior of a single particle in RDPSO are proved in the Appendix. As shown by Theorem A2, the particle's behavior is related to the convergence of $\rho_n = \prod_{i=1}^{n} \lambda_i$, where $\lambda_n = \alpha\varphi_n + (1-\beta)$ subject to a normal distribution, namely, $\lambda_n \sim N(1-\beta, \alpha^2)$. It is concluded by Theorem A3 that if and only if $\Delta = E(\ln|\lambda_n|) \le 0$, namely, the values of $\alpha$ and $\beta$ satisfy the following relationship:

$$\Delta = \frac{1}{\sqrt{2\pi}\alpha}\int_{-\infty}^{+\infty} \ln|x|\, e^{-\frac{[x-(1-\beta)]^2}{2\alpha^2}}\, dx \le 0\,, \tag{20}$$

$\rho_n$ is probabilistically bounded and, thus, the position of the particle is probabilistically bounded too. In inequality (20), the value of $\Delta$ is an improper integral which is undefined at $x=0$. By Dirichlet test, this improper integral is convergent if both $\alpha$ and $\beta$ are two finite numbers (Courant, 1989).

Inequality (20) does not provide any explicit constraint relation between $\alpha$ and $\beta$ due to the difficulty in calculating the improper integral in the inequality. A sufficient condition for $\Delta < 0$ (i.e. $\lim_{n\to\infty}\rho_n = \lim_{n\to\infty}\prod_{i=1}^{n}\lambda_i = 0$) is derived in Theorems A4. It says that if the values of $\alpha$ and $\beta$ are subject to the constraint:

$$0 < \alpha < 1,\quad 0 < \beta < 2, \tag{21}$$

$\Delta < 0$ and $\rho_n = \prod_{i=1}^{n}\lambda_i$ converges to zero, which consequently ensures the probabilistic boundedness of the particle's position as shown. Figure 1 visualizes some simulation results on the stochastic behaviour of the particle by using different values of $\alpha$ and $\beta$, with $C$ fixed at $X = 0.001$, $p$ fixed at the origin and the initial position of the particle set as $X_0 = 1000$. Figures 1 (a) to (c) show the results with $\alpha$ and $\beta$ satisfying constraint (21). It can be observed that the particle's position oscillated around $p$ and $C$, implying that the position is probabilistically bounded in these cases. Figures 1 (d) to (i) show that the particle's position is probabilistically bounded in some cases when $\alpha$ and $\beta$ do not satisfy constraint (21). This verifies that constraint (21) is a sufficient condition for $\Delta < 0$ or $\lim_{n\to\infty}\rho_n = 0$. At other values of $\alpha$ and $\beta$ not satisfying (21), the value of $\ln|X_n - p|$ reached 700 and stopped changing after a certain number of iterations, as shown in Figures 1 (j) to (o). In such cases, the value of $|X_n - p|$ reaches the maximum positive value that the computer can identify, so that it can be considered to have diverged to infinity.

Constraint (21) is of practical significance to the application of the RDPSO algorithm, although it does not give the necessary condition for $\Delta \le 0$. In practice, the values of $\alpha$ and $\beta$ can generally be selected within the intervals given by (21), for a satisfactory algorithmic performance when the algorithm is applied to real-world problems. In Section 4, a detailed investigation into how to select these algorithmic parameters is undertaken by using a set of benchmark functions from the CEC2005 benchmark suite.

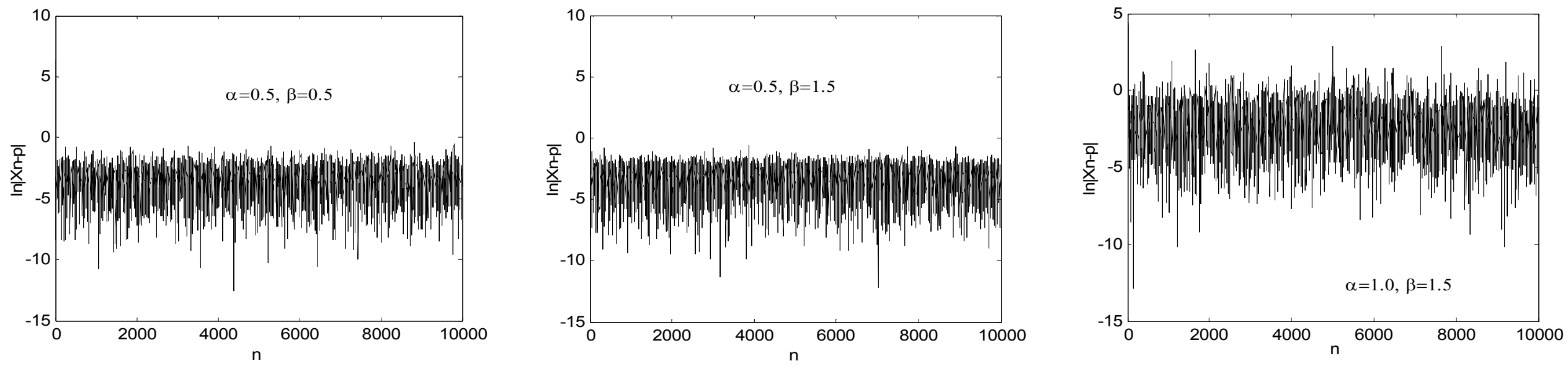

(a) (b) (c)

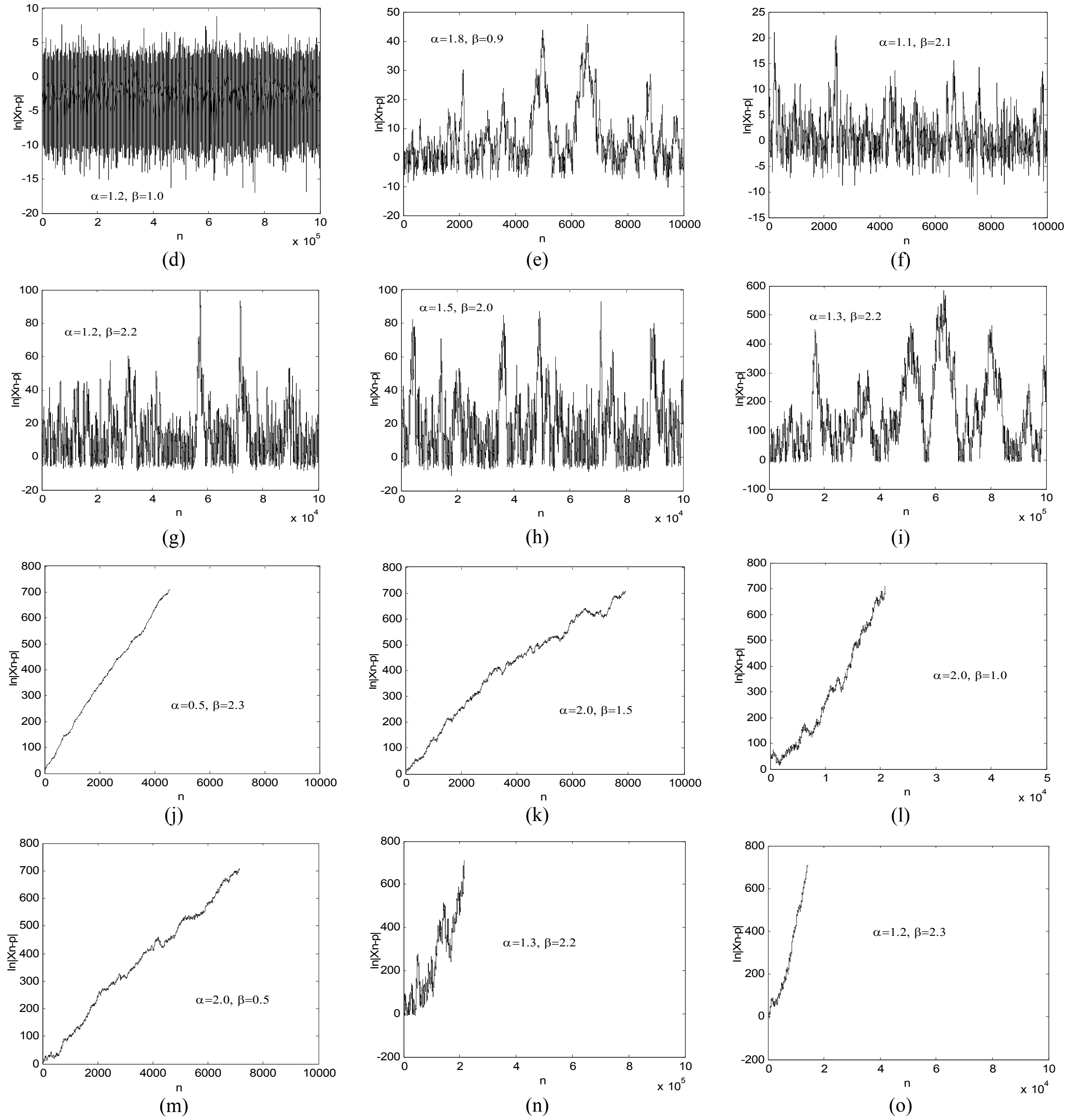

Figure 1 The figure visualizes the simulation results for the behavior of the particle at different values of $\alpha$ and $\beta$. Figures (a) to (c) show that when the values of $\alpha$ and $\beta$ are selected within the intervals $(0,1)$ and $(0,2)$, the particle's position is probabilistically bounded. Figures (d) to (i) show that the particle's position may be also probabilitcally bounded at some values of $\alpha$ and $\beta$ not satisfying constraint (21). Figures (j) to (o) show some cases that when $\alpha$ and $\beta$ do not satisfy constraint (21), $\ln|X_n - p| \to +\infty$ (i.e. $|X_n - p| \to +\infty$) as *n* increases.

### 3.3 The RDPSO's Search Behavior

In the above analysis, it is assumed that each particle in the RDPSO updates its velocity and position independently, with the mean best position $C$ and the local focus $p$ being treated as independent probabilistically bounded random variables, and thus it is revealed that the behavior of the particle is related to the convergence or the boundedness of $\rho_n$. However, the actual situation is more complex when the RDPSO algorithm is running in a real-world landscape. During the search process of the RDPSO algorithm, each particle is influenced not only by $\rho_n$ but also by the points $C_n$ and $p_{i,n}$, which can not be treated as independent random variables anymore, but are relevant to the other particles. As for $C_n$, it is the mean of the *pbest* positions of all the particles, moving with the variation of each *pbest* position. The local focus $p_{i,n}$, is a random point associated with the *pbest* position of particle *i* ($P_{i,n}$) and the *gbest* position $G_n$ that rotates among the *pbest* positions of the member particles according to their fitness values. In contrast to $C_n$, $p_{i,n}$, as well as $P_{i,n}$ and $G_n$, varies more dramatically, since $C_n$ averages the changes of all the *pbest* positions.

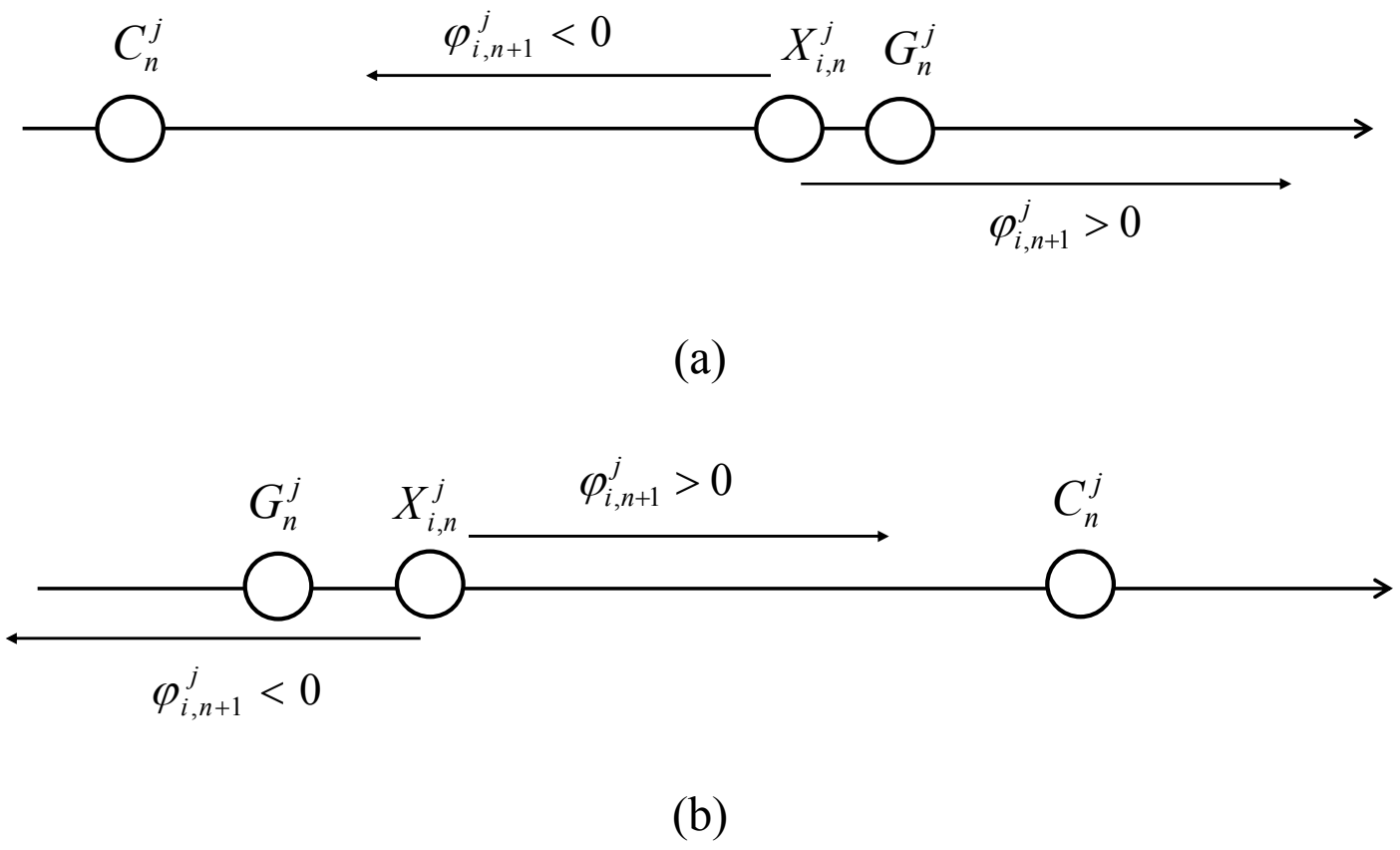

Figure 2 The figure shows that the *mbest* position $C_n^j$ pulls or pushes the particle away from $G_n^j$. The direction of the particle's movement is determined by the sign of $\varphi_{i,n+1}^j$

Generally, the *pbest* positions of all the particles converge to a single point when the RDPSO algorithm is performing an optimization task, which implies that $P\{\lim_{n\to\infty} |C_n - p_{i,n}| = 0\} = 1$ as mentioned in the proof of Theorem A1. Referring to equations (A7) to (A10), we can infer that if and only if $\Delta < 0$, $\lim_{n\to\infty} |X_{i,n} - C_n| = 0$ or $\lim_{n\to\infty} |X_{i,n} - p_{i,n}| = 0$. That means the current positions and the *pbest* positions of all the particles converge to a single point when $\Delta < 0$. It can also be found from Theorems A2 and A3 that, when $\Delta = 0$, the particle's position is probabilistically bounded and oscillates around but does not converge to $C_n$ or $p_{i,n}$, even though $P\{\lim_{n\to\infty} |C_n - p_{i,n}| = 0\} = 1$. When $\Delta > 0$, it is shown by Theorems A2 and A3 that the particle's current position diverges and the explosion of the whole particle swarm happens.

In practical applications, it is always expected that the particle swarm in the RDPSO algorithm can converge to a single point, like that in the canonical PSO. Essentially, there are two movement trends, i.e. the random motion and the drift motion, for each particle in the RDPSO, as has been described in the motivation of the algorithm. These two motions reflect the global search and the local search, respectively. The drift motion, represented by the $VD_{i,n+1}^{j}$ in the velocity update equation (15), draws the particle towards the local focus and makes the particle search in the vicinity of the *gbest* position and its *pbest* position so that the particle's current and *pbest* positions can constantly come close to the *gbest* position. On the other hand, the random component $VR_{i,n+1}^{j}$ results in a random motion, leading the particle to be so volatile that its current position may reach a point far from the *gbest* position and its *pbest* position. This component can certainly provide the particle a global search ability, which, in the canonical PSO algorithm, is given by the velocity at the last iteration, i.e. $V_{i,n+1}^{j}$. Nevertheless, an important characteristic distinguishing the RDPSO from other randomized PSO methods is that the random component of the particle's velocity uses an adaptive standard deviation for its distribution, i.e. $\alpha | X_{i,n}^{j} - C_{n}^{j} |$. Such a random component makes the random motion of the particle have a certain orientation. The effect of $VR_{i,n+1}^{j}$ is to pull or push the particle away from the *gbest* position by $C_{n}^{j}$ as shown by Figure 2, not only to displace the particle randomly as the mutation operation does in some variants of PSO and evolutionary algorithms. Figure 2(a) shows that, when $C_{n}^{j}$ is at the left side of $X_{i,n}^{j}$ and $G_{n}^{j}$, $| X_{i,n}^{j} - C_{n}^{j} | = X_{i,n}^{j} - C_{n}^{j}$. The drift component $\beta(p_{i,n}^{j} - X_{i,n}^{j})$ draws the particle right towards $G_{n}^{j}$. If $\varphi_{i,n+1}^{j} > 0$, $\alpha | X_{i,n}^{j} - C_{n}^{j} | \varphi_{i,n+1}^{j} = \alpha(X_{i,n}^{j} - C_{n}^{j})\varphi_{i,n+1}^{j} > 0$, which makes the particle move to the right further and, thus, pushes $X_{i,n}^{j}$ away from $G_{n}^{j}$. If $\varphi_{i,n+1}^{j} < 0$, $\alpha(X_{i,n}^{j} - C_{n}^{j})\varphi_{i,n+1}^{j} < 0$, whose effect is that the particle's position is pulled away from $G_{n}^{j}$. Figure 2(b) illustrates the case when $C_{n}^{j}$ is at the right side of $X_{i,n}^{j}$ and $G_{n}^{j}$. Only the effect of the sign of $\varphi_{i,n+1}^{j}$ on the direction of the particle's motion is opposite to that in Figure 2(a). Generally speaking, the longer the distance $| X_{i,n}^{j} - C_{n}^{j} |$, the farther the particle's position at next iteration $X_{i,n+1}^{j}$ will be away from the *gbest* position. If the particle's position is close to the *gbest* position, the random component can help the particle escape the *gbest* position easily, when the *gbest* position is stuck into a local optimal solution. As far as the whole particle swarm is concerned, the overall effect is that the RDPSO has a better balance between the global search and the local search, as illustrated below.

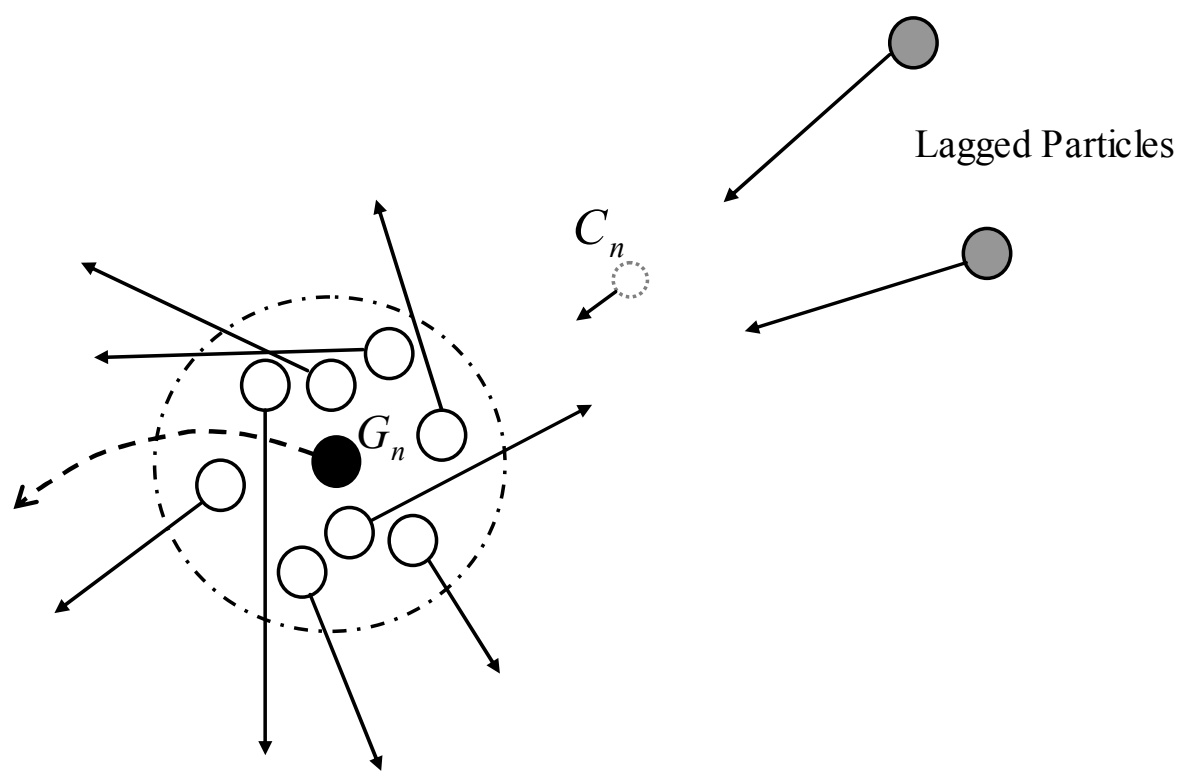

Figure3. The figure shows that $C_n$ is shifted toward the lagged particles and thus far from the particles clustering around $G_n$. The particles are pulled or pushed away from the neighbourhood of $G_n$ and would search the landscape globally.

In the RDPSO method, the swarm could not gather around the *gbest* position without waiting for the lagged particles. Figure 3 depicts the concept where the *pbest* positions of several particles, known as the lagged particles, are located far away from the rest of the particles and the *gbest* position $G_n$, while the rest of the particles are nearer to the global best position, with their *pbest* positions located within a neighbourhood of the *gbest* position. The *mbest* position $C_n$ would be shifted towards the lagged particles and be located outside the neighbourhood. When the lagged particles are chasing after their colleagues, that is, converging to $G_n$, $C_n$ is approaching $G_n$ slowly. The current positions of the particles within the neighbourhood would be pulled or pushed outside the neighbourhood by $C_n$, and the particles would explore the landscape globally around $G_n$ so that the current $G_n$ could skip out onto a better solution. As $C_n$ is careening toward the neighbourhood, the exploration scope of the particle is becoming narrower. After the lagged particles move into the neighbourhood of the *gbest* position, $C_n$ also enter the neighbourhood and the particles would perform the same search process based on a smaller neighbourhood of the *gbest* position. In the canonical PSO, each particle converges to the *gbest* position independently and has less opportunity to escape from the neighbourhood of the *gbest* position. When the speed of the particle is small, it is impossible for the particles within the neighbourhood to jump out of the neighbourhood. As a result, these particles would perform local search around the *gbest* position and only the lagged particles could search globally. Evident from the above analysis, the RDPSO algorithm generally has a better balance between exploration and exploitation than the canonical PSO.

Moreover, different from mutation operations that play minor roles in some variants of PSO and evolutionary algorithms, the random motion has an equally important role as the drift motion in the RDPSO. Owing to the random motion oriented by $C_n$, the RDPSO achieves a good balance between the local and global searches during the search process. By the influences of both $C_n$ and their local focuses, the particles in the RDPSO have two movement trends, convergence and divergence, but the overall effect is their convergence to a common point of all the particles if $\Delta < 0$. The convergence rate of the algorithm depends on the values of $\alpha$ and $\beta$, which can be tuned to balance the local and global search, when the algorithm is used for a practical problem.

### 3.4. Variants of RDPSO

In order to investigate the RDPSO in depth, some variants of the algorithm are proposed in this paper. Two methods are used for determining the random component of the velocity. One employs equation (13) for this component and the other replaces the *mbest* position in (13) by the *pbest* position of a randomly selected particle in the population at each iteration. For convenience, we denote the randomly selected *pbest* position by $C'_n$. For each particle, the probability for its *pbest* position to be selected as $C'_n$ is 1/*M*. Consequently, the expected value of $C'_n$ equals to $C_n$, that is,

$$E(C'_n) = \sum_{i=1}^{M} \frac{1}{M} P_{i,n} = C_n \,. \tag{22}$$

However, since the $C'_n$ appears to be more changeful than $C_n$, the current position of each particle at each iteration shows to be more volatile than that of the particle with equation (13), which diversifies the particle swarm and in turn enhances the global search ability of the algorithm.

In addition to the global best model, the local best model is also examined for the RDPSO. The ring topology is a widely used neighborhood topology for the local best model (Li, 2010), in which each particle connects exactly to two neighbors. The standard PSO (SPSO) in (Bratton and Kennedy, 2007) is defined by the integration of the PSO-Co with the ring topology. Although there are various neighborhood topologies, we chose the ring topology for the RDPSO with the local best model. Thus, the combination of the two topologies with the two strategies for the random velocity component produces the four resulting RDPSO variations:

**RDPSO-Gbest**: The RDPSO algorithm with the global best model and the random velocity component described by equation (13).

**RDPSO-Gbest-RP**: The RDPSO algorithm using the global best model and employing a randomly selected *pbest* position to determine the random velocity component.

**RDPSO-Lbest**: The RDPSO algorithm with the ring neighborhood topology and the random velocity component in (13), where, however, the *mbest* position is the mean of the *pbest* positions of the neighbors of each particle and the particle itself, instead of the mean of the *pbest* positions of all the particles in the population.

**RDPSO-Lbest-RP**: The RDPSO algorithm using the ring neighborhood topology and employing the *pbest* position of a particle randomly selected from the neighbors of each particle and the particle itself.

# 4. Experimental Results and Discussion

## 4.1. Benchmark Problems

The previous analysis of the RDPSO provides us with a deep insight into the mechanism of the algorithm. However, it is not sufficient to evaluate the effectiveness of the algorithm without comparing it with other methods on a set of benchmark problems. To evaluate the RDPSO in an empirical manner, the twenty five functions from the CEC2005 benchmark suite (Suganthan, 2005) were employed for this purpose. Functions $F_1$ to $F_5$ are unimodal, functions $F_6$ to $F_{12}$ are multi-modal, $F_{13}$ and $F_{14}$ are two expanded functions, and $F_{15}$ to $F_{25}$ are hybrid functions. The dimension $N$ was chosen as 30 for each of these functions. The mathematical expressions and properties of the functions are described in detail in (Suganthan, 2005). The codes in Matlab, C and Java for the functions could be found at http://www.ntu.edu.sg/home/EPNSugan/.

## 4.2. Empirical Studies on the Parameter Selection of the RDPSO Variants

Parameter selection is the major concern when a stochastic optimization algorithm is being employed to solve a given problem. For the RDPSO, the algorithmic parameters include the population size, the maximum number of iterations, the thermal coefficient $\alpha$ and the drift coefficient $\beta$. Like in the canonical PSO, the population size in the RDPSO is recommended to be set from 20 to 100. The selection of the maximum number of iterations depends on the problem to be solved. In the canonical PSO, the acceleration coefficients and the inertia weight (or the constriction factor) have been studied extensively and in depth since these parameters are very important for the convergence of the algorithm. For the RDPSO algorithm, $\alpha$ and $\beta$ play the same roles as the inertia weight and the acceleration coefficients for the canonical PSO. In Section 3, it was shown that it is sufficient to set $\alpha$ and $\beta$ according to (21), such that $\Delta < 0$, to prevent the individual particle from divergence and guarantee the convergence of the particle swarm. However, this does not mean that such values of $\alpha$ and $\beta$ can lead to a satisfactory performance of the RDPSO algorithm in practical applications. This section intends to find out, through empirical studies, suitable settings of $\alpha$ and $\beta$ so that the RDPSO may yield good performance in general. It also aims to compare the performance of the RDPSO variants with other forms of PSO.

There are various control methods for the parameters $\alpha$ and $\beta$ when the RDPSO is applied to practical problems. A simple approach is to set them as fixed values when the algorithm is executed. Another method is to decrease the value of the parameter linearly during the course of the search process. In this work, we fixed the value of $\beta$ in all the experiments and employed the two control methods for $\alpha$, respectively.

To specify the value of $\alpha$ and $\beta$ for real applications of the RDPSO, two groups of experiments were performed on the benchmark functions. The first group of experiments tested the RDPSO algorithm with different parameter settings on several benchmark functions in order to find out the parameter settings that can yield satisfactory performance on these functions. Then, the identified parameter settings were tested by more benchmark functions, to verify their generalizability, in the second group of experiments.

In the first set of experiments, we tested the RDPSO-Gbest, RDPSO-Gbest-RP, RDPSO-Lbest, and RDPSO-Lbest-RP on three frequently used functions in the CEC2005 benchmark suite: Shifted Rosenbrock Function ($F_6$), Shifted Rotated Griewank's Function ($F_7$), and Shifted Rastrigin's Function ($F_9$), using the two methods for controlling $\alpha$ with $\beta$ fixed at 1.5 or 1.45. The initial position of each particle was determined randomly within the initialization range. For each parameter configuration, each algorithm, using 40 particles, was tested for 100 runs on each problem. To determine the effectiveness of each algorithm for the $\alpha$ setting under each control method with a fixed value of $\beta$ on each problem, the best objective function value (i.e., the best fitness value) found after 5000 iterations was averaged over 100 runs of tests for that parameter setting and the same benchmark function. The results (i.e., the mean best fitness values) obtained by the parameter settings with the same control method for $\alpha$ were compared across the three benchmarks. The best parameter setting with each control method for $\alpha$ was selected by ranking the averaged best objective function values for each problem, summing the ranks, and taking the value that had the lowest summed (or average) rank, provided that the performance is acceptable (in the top half of the rankings) in all the tests for a particular parameter configuration.

The rankings of the results for the RDPSO-Gbest are plotted in Figure 4. When the fixed value method was used for $\alpha$, it was set to a range of values subject to constraint (21), with $\beta$ setting at 1.5 or 1.45 in each case. Results obtained for other parameter settings were very poor and are not considered for ranking. The best average rank among all the tested parameter configurations occurs when $\alpha = 0.7$ and $\beta = 1.5$. When linearly varying $\alpha$ was used, its initial value $\alpha_1$ and final value $\alpha_2$ ($\alpha_1 > \alpha_2$) were selected from a series of different values subject to constraint (21), with $\beta$ setting at 1.5 or 1.45. Only acceptable results are ranked and plotted in Figure 4. It was found that with $\beta = 1.45$, decreasing $\alpha$ linearly from 0.9 to 0.3 leads to the best performance among all the tested parameter settings.

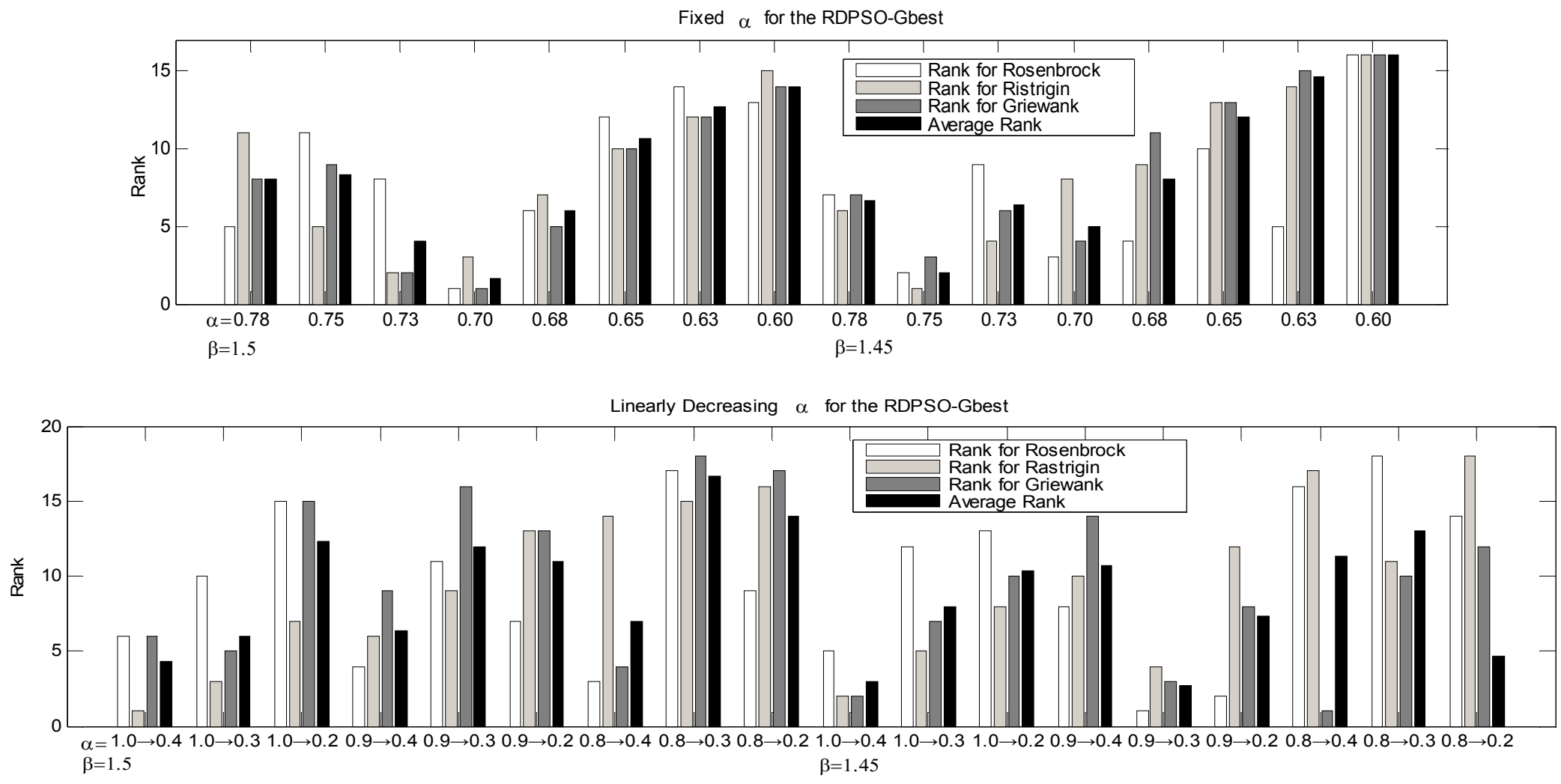

Figure 4. The rankings of the mean best fitness values for each of the three benchmarks and the average rank for the RDPSO-Gbest.

The rankings of the results for the RDPSO-Gbest-RP are visualized in Figure 5. It is clear from these results that the value of $\alpha$, whether it used the fixed value or time-varying method, should be set relatively small, so that the algorithm is comparable in performance with the RDPSO-Gbest, when $\beta$ was given. Results obtained with $\alpha$ outside the range [0.38, 0.58] were of poor quality and were not used for ranking. As shown in Figure 5, when the fixed value method for $\alpha$ was used, the best average ranks among all tested parameter settings were obtained by setting $\alpha = 0.5$ and $\beta = 1.45$. On the other hand, the algorithm exhibited the average best performance when $\beta = 1.45$ and $\alpha$ was decreasing linearly from 0.6 to 0.2, for the method of linearly varying $\alpha$.

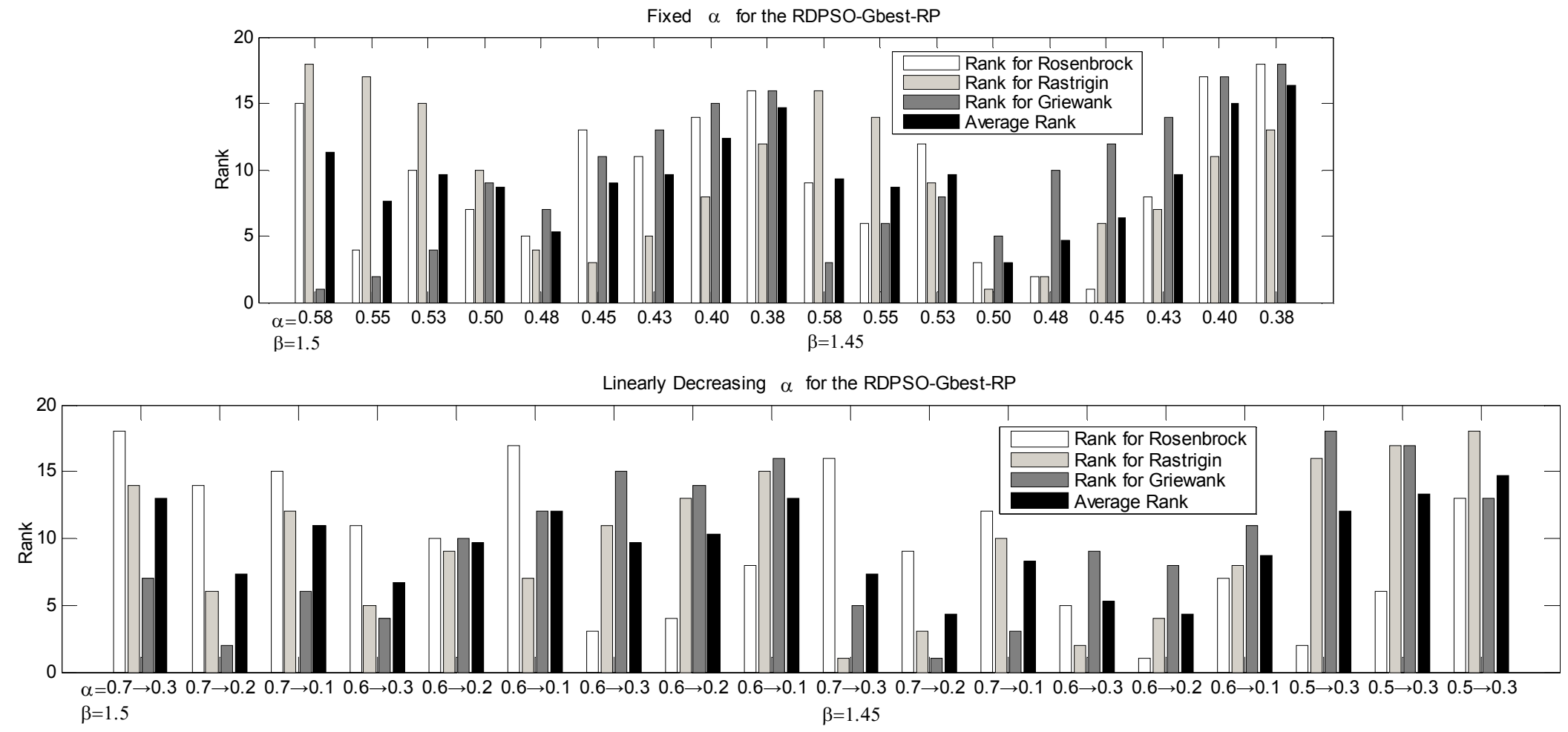

Figure 5. The rankings of the mean best fitness values for each of the three benchmarks and the average rank for the RDPSO-Gbest-RP.

Figure 6 shows the rankings of the results for the RDPSO-Lbest. For the fixed $\alpha$ method, the results of the algorithm obtained with $\alpha$ outside the range [0.6, 0.78] did not participate in rankings because of their poor qualities. The best average rank among all the tested parameter configurations in this case occur when $\alpha = 0.7$ and $\beta = 1.5$. For the linearly varying $\alpha$ method, it was identified that varying $\alpha$ linearly from 0.9 to 0.3 with $\beta = 1.45$ could yield the average best quality results among all the tested parameter configurations.

Figure 7 plots the rankings of the results for the RDPSO-Lbest-RP. For fixed $\alpha$, the best average rank among all the tested parameter settings could be obtained when $\alpha = 0.7$ and $\beta = 1.45$. For time-varying $\alpha$, the algorithm obtained the average best performance among all the tested parameter configurations when $\alpha$ was decreasing linearly from 0.9 to 0.3, with $\beta = 1.45$.

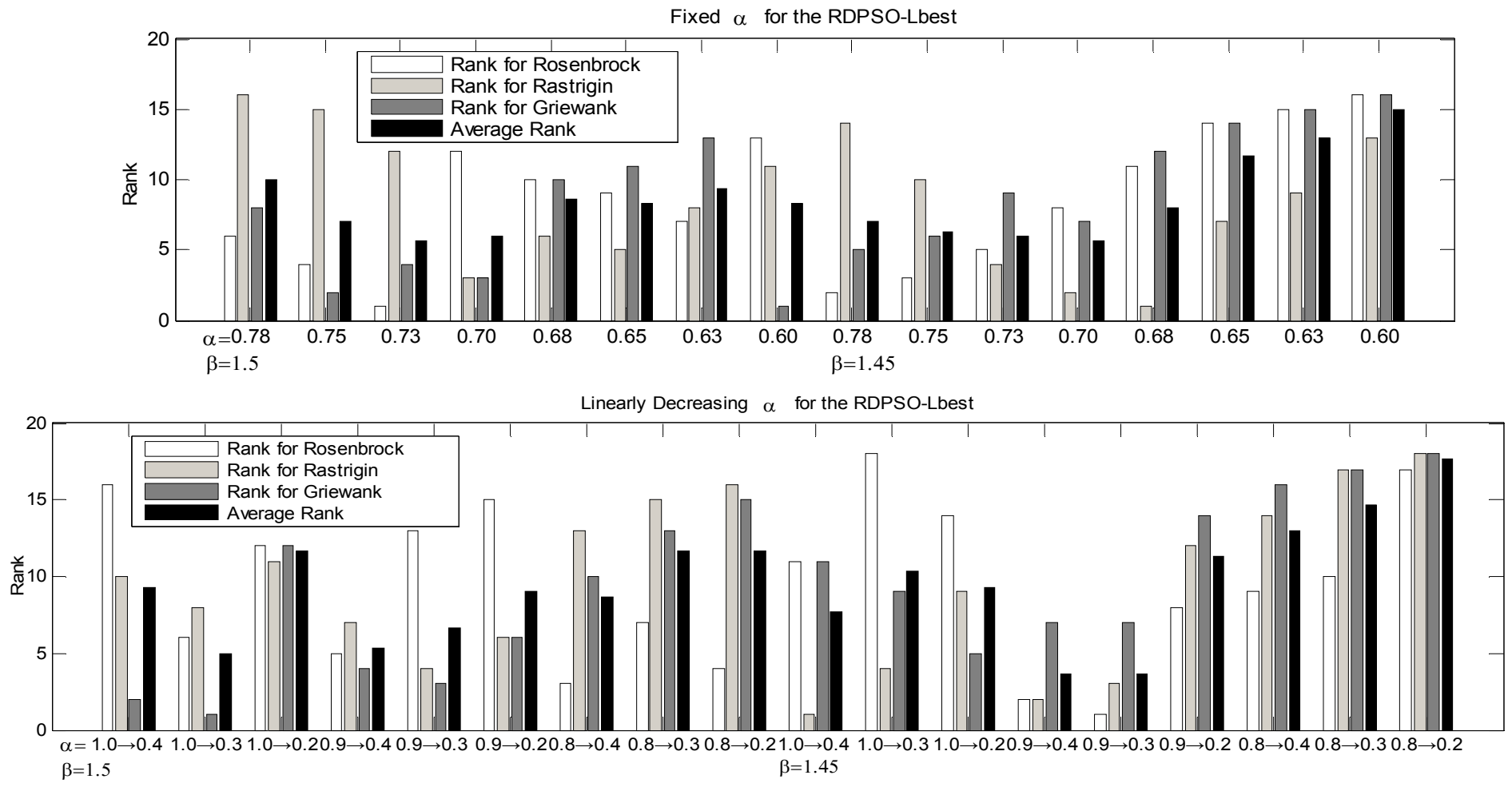

Figure 6. The rankings of the mean best fitness values for each of the three Benchmarks and the average rank for the RDPSO-Lbest.

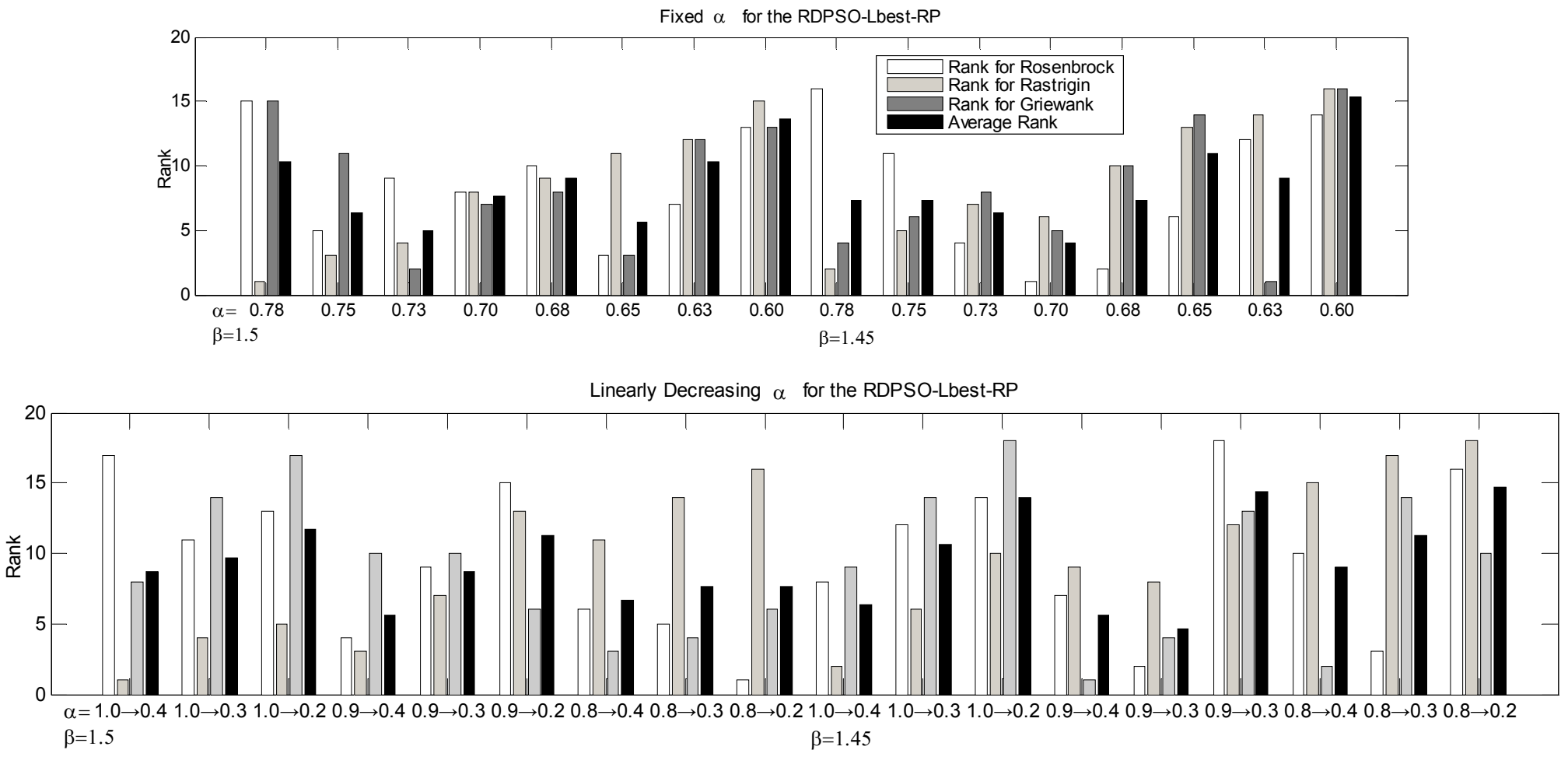

Figure 7. The rankings of the mean best fitness values for each of the three Benchmarks and the average rank for the RDPSO-Lbest-RP.

In the second group of experiments, each variant of the RDPSO with the parameter settings for $\alpha$ and $\beta$ that were identified to yield better performance on average in the first group of experiments were tested by the first twelve benchmark functions in the CEC2005 benchmark suite. The purpose here is to find out whether the identified parameter settings are able to obtained satisfactory performance for other problems and to make a further performance comparison between the two methods of controlling $\alpha$ for each RDPSO variant. The results listed in Table 1 include the means and standard deviations of the best fitness values obtained over 100 runs of each RDPSO variant on each of the twelve benchmarks. For the RDPSO-Gbest, the linearly varying $\alpha$ method outperformed the fixed $\alpha$ method on all the twelve functions except $F_4$ and $F_{11}$. For the RDPSO-Gbest-RP, the linearly varying $\alpha$ generated better results than the fixed $\alpha$ on eight functions (i.e. $F_3$, $F_5$, $F_6$, $F_7$, $F_9$, $F_{10}$, $F_{11}$ and $F_{12}$). It can be seen that the linearly varying $\alpha$ method outperformed the fixed $\alpha$ method on all of the twelve functions for the RDPSO-Lbest. For the RDPSO-Lbest-RP, the linearly varying $\alpha$ had better performance than the fixed $\alpha$ on five benchmarks. On the other benchmark functions, the linearly varying $\alpha$ method did not show significant poorer performance compared to the fixed $\alpha$ method. However, among the eight versions of the RDPSO, the total rank of the RDPSO-RP with the linearly varying $\alpha$ are smaller than that with the fixed $\alpha$. Besides, in the first set of experiments, we found that the performance of each RDPSO variant with the fixed $\alpha$ is more sensitive to the value of $\alpha$ than that with the linearly varying $\alpha$. Considering the above facts, we recommended that the linearly varying $\alpha$ method should be used for real applications.

Table 1. Performance Comparison Among the RDPSO Variants with Different Parameter Control Strategies for $F_1$ to $F_{12}$

| Algorithms | $F_1$ | $F_2$ | $F_3$ | $F_4$ | $F_5$ | $F_6$ |
|---|---|---|---|---|---|---|
| RDPSO-Gbest ($\alpha$=0.7, $\beta$=1.5) | 6.0670e-027 (1.0497e-027) | 0.9648 (2.8121) | 6.7963e+006 (4.2190e+006) | 289.2492 (296.1889) | 3.6992e+003 (1.1087e+003) | 62.8787 (84.2366) |
| RDPSO-Gbest ($\alpha$=0.9→0.3,$\beta$=1.45) | 2.2871e-027 (4.3476e-028) | **0.0805 (0.1341)** | 4.7079e+006 (3.1653e+006) | 411.2758 (574.1945) | 2.6293e+003 (808.8539) | 60.9164 (78.5198) |
| RDPSO-Gbest-RP ($\alpha$=0.5, $\beta$=1.45) | 5.3927e-037 (7.2863e-038) | 0.0107 (0.0223) | 4.6152e+006 (2.5412e+006) | **53.4919 (59.6771)** | 3.2702e+003 (1.1447e+003) | 45.3846 (51.7045) |
| RDPSO-Gbest-RP ($\alpha$=0.6→0.2,$\beta$=1.45) | 8.1256e-037 (1.4983e-037) | 0.1131 (1.0156) | **2.5203e+006 (1.6334e+006)** | 217.8821 (269.0046) | **2.2241e+003 (865.3596)** | 34.9274 (39.0403) |
| RDPSO-Lbest ($\alpha$=0.7, $\beta$=1.45) | 8.5371e-029 (2.6319e-029) | 2.4408 (2.3839) | 6.5290e+006 (2.9835e+006) | 2.4422e+003 (1.1875e+003) | 3.4503e+003 (774.0588) | 41.3477 (49.0390) |
| RDPSO-Lbest ($\alpha$=0.9→0.3,$\beta$=1.45) | 3.9443e-031 (9.5470e-031) | 2.4034 (1.7191) | 4.9772e+006 (1.9029e+006) | 1.6199e+003 (883.4518) | 2.7654e+003 (638.3375) | **19.5009 (16.7704)** |
| RDPSO-Lbest-RP ($\alpha$=0.7, $\beta$=1.45) | 2.8636e-030 (2.1470e-030) | 9.3696 (5.6907) | 3.4615e+006 (1.2020e+006) | 3.8994e+003 (1.6189e+003) | 4.4133e+003 (1.0664e+003) | 21.5193 (30.8859) |
| RDPSO-Lbest-RP ($\alpha$=0.9→0.3,$\beta$=1.45) | **5.2461e-037 (7.3587e-038)** | 9.3880 (6.7340) | 4.8092e+006 (1.7477e+006) | 3.4502e+003 (1.3764e+003) | 3.9088e+003 (888.7718) | 24.0065 (24.4861) |

| Algorithms | $F_7$ | $F_8$ | $F_9$ | $F_{10}$ | $F_{11}$ | $F_{12}$ |
|---|---|---|---|---|---|---|
| RDPSO-Gbest ($\alpha$=0.7, $\beta$=1.5) | 0.0200 (0.0173) | 20.9583 (0.0601) | 45.7970 (15.8581) | 160.1897 (51.7502) | **15.8921 (6.1393)** | 8.3755e+003 (6.8869e+003) |
| RDPSO-Gbest ($\alpha$=0.9→0.3,$\beta$=1.45) | 0.0175 (0.0140) | 20.9558 (0.0641) | **22.7650 (5.7728)** | 78.6024 (38.9282) | 21.6689 (7.6173) | 6.0361e+003 (5.1260e+003) |
| RDPSO-Gbest-RP ($\alpha$=0.5, $\beta$=1.45) | 0.0165 (0.0176) | 20.9579 (0.0497) | 37.0594 (14.8284) | 198.0320 (16.9197) | 20.8469 (12.4781) | 3.0143e+003 (4.1409e+003) |
| RDPSO-Gbest-RP ($\alpha$=0.6→0.2,$\beta$=1.45) | 0.0130 (0.0123) | 20.9602 (0.0569) | 31.9085 (8.7969) | 82.5152 (47.7362) | 20.0701 (6.9879) | **2.8227e+003 (3.3963e+003)** |
| RDPSO-Lbest ($\alpha$=0.7, $\beta$=1.45) | 0.0108 (0.0077) | 20.9650 (0.0466) | 41.7659 (13.4442) | 146.1542 (36.9750) | 25.8698 (3.7851) | 7.9082e+003 (5.5806e+003) |
| RDPSO-Lbest ($\alpha$=0.9→0.3,$\beta$=1.45) | 0.0092 (0.0050) | **20.9540 (0.0508)** | 26.9390 (6.0386) | **49.8606 (12.9486)** | 22.1984 (3.0396) | 4.0616e+003 (2.8103e+003) |
| RDPSO-Lbest-RP ($\alpha$=0.7, $\beta$=1.45) | **0.0092 (0.0052)** | 20.9561 (0.0557) | 47.2217 (11.8387) | 68.0621 (17.2223) | 22.4425 (2.0700) | 3.1786e+003 (2.4794e+003) |
| RDPSO-Lbest-RP ($\alpha$=0.9→0.3,$\beta$=1.45) | 0.0093 (0.0061) | 20.9613 (0.0543) | 36.4589 (7.9391) | 51.4390 (7.9391) | 23.0731 (1.8929) | 3.7315e+003 (1.9675e+003) |

### 4.3. Performance Comparisons among the RDPSO Variants and Other PSO Variants

To investigate the generalizability of the parameter settings for the linearly varying $\alpha$ method used in the RDPSO, and to the determine whether RDPSO can be as effective as other variants of PSO, a further performance comparison using all the twenty five benchmark functions of the CEC2005 benchmark suite was made between the RDPSO algorithms (i.e., the RDPSO-Gbest, RDPSO-Gbest-RP, RDPSO-Lbest and RDPSO-Lbest-RP) and other variants of PSO, including the PSO with inertia weight (PSO-In) (Shi and Eberhart, 1998a; 1998b; 1999), the PSO with constriction factor (PSO-Co) (Clerc and Kennedy, 2002; Clerc 1999), the PSO-In with local best model (PSO-In-Lbest) (Liang et al., 2006), the standard PSO (SPSO) (i.e. PSO-Co-Lbest) (Bratton and Kennedy, 2007), the Gaussian bare bones PSO (GBBPSO) (Kennedy, 2003; 2004), the comprehensive learning PSO (CLPSO) (Liang et al., 2006), the dynamic multiple swarm PSO (DMS-PSO) (Liang and Suganthan, 2005), and the fully-informed particle swarm (FIPS) (Mendes, 2004). Each algorithm was run 100 times on each problem using 40 particles to search the global optimal fitness value. At each run, the particles in the algorithms started in new and randomly-generated positions, which are uniformly distributed within the search bounds. Each run of each algorithm lasted 5000 iterations, and the best fitness value (objective function value) for each run was recorded.

Table 2. Mean and Standard Deviation of the Best Fitness Values after 100 runs of Each Algorithm for $F_1$ to $F_6$

| Algorithms | $F_1$ | $F_2$ | $F_3$ | $F_4$ | $F_5$ | $F_6$ |
|---|---|---|---|---|---|---|
| PSO-In | 3.9971e-028<br>(5.6544e-028) | 263.2219<br>(608.4657) | 3.4324e+007<br>(3.0220e+007) | 2.7829e+003<br>(2.0996e+003) | 4.3961e+003<br>(1.5331e+003) | 143.7144<br>(336.9297) |
| PSO-Co | 6.7053e-029<br>(1.0671e-028) | **0.0100**<br>**(0.0939)** | 1.3659e+007<br>(1.3662e+007) | 842.4768<br>(1.5264e+003) | 6.2857e+003<br>1.9629e+003 | 57.5740<br>(84.2278) |
| PSO-In-Lbest | 2.7049e-013<br>(5.1148e-013) | 865.7861<br>(368.9980) | 2.5658e+007<br>(1.0089e+007) | 8.7648e+003<br>(1.8468e+003) | 8.0095e+003<br>(1.0568e+003) | 57.5362<br>(74.5821) |
| SPSO<br>(PSO-Co-Lbest) | 4.2657e-036<br>(2.3958e-036) | 0.8615<br>(0.7092) | 3.3604e+006<br>(1.5549e+006) | 6.3348e+003<br>(2.3147e+003) | 5.2549e+003<br>(1.1583e+003) | 47.3744<br>(79.8406) |
| GBBPSO | 7.0941e-027<br>(1.9421e-026) | 0.0110<br>(0.0174) | 4.9003e+006<br>(2.6581e+006) | 1.0432e+003<br>1.0819e+003 | 8.0391e+003<br>(2.8824e+003) | 109.8415<br>(330.4848) |
| FIPS | 1.2395e-036<br>(8.4958e-037) | 0.1390<br>(0.0682) | 6.9970e+006<br>(2.4490e+006) | 4.5429e+003<br>(1.4685e+003) | 3.3929e+003<br>(599.5893) | 109.1170<br>(179.8489) |
| DMS-PSO | 8.8399e-016<br>(2.1311e-015) | 141.1109<br>(70.6632) | 5.6008e+006<br>(2.9187e+006) | 976.6745<br>(391.0695) | 2.4263e+003<br>(498.7101) | 211.0941<br>(314.9179) |
| CLPSO | 5.2323e-017<br>(2.9219e-017) | 1.2661e+003<br>(297.3666) | 3.3326e+007<br>(8.8808e+006) | 7.6045e+003<br>(1.7722e+003) | 4.0357e+003<br>(489.0741) | 74.2914<br>(31.5737) |
| RDPSO-Gbest | 2.2871e-027<br>(4.3476e-028) | 0.0805<br>(0.1341) | 4.7079e+006<br>(3.1653e+006) | 411.2758<br>(574.1945) | 2.6293e+003<br>(808.8539) | 60.9164<br>(78.5198) |
| RDPSO-Gbest-RP | 8.1256e-037<br>(1.4983e-037) | 0.1131<br>(1.0156) | **2.5203e+006**<br>**(1.6334e+006)** | **217.8821**<br>**(269.0046)** | **2.2241e+003**<br>**(865.3596)** | 34.9274<br>(39.0403) |
| RDPSO-Lbest | 3.9443e-031<br>(9.5470e-031) | 2.4034<br>(1.7191) | 4.9772e+006<br>(1.9029e+006) | 1.6199e+003<br>(883.4518) | 2.7654e+003<br>(638.3375) | **19.5009**<br>**(16.7704)** |
| RDPSO-Lbest-RP | **5.2461e-037**<br>**(7.3587e-038)** | 9.3880<br>(6.7340) | 4.8092e+006<br>(1.7477e+006) | 3.4502e+003<br>(1.3764e+003) | 3.9088e+003<br>(888.7718) | 24.0065<br>(24.4861) |

Table 3. Mean and Standard Deviation of the Best Fitness Values after 100 runs of Each Algorithm for $F_7$ to $F_{12}$

| Algorithms | $F_7$ | $F_8$ | $F_9$ | $F_{10}$ | $F_{11}$ | $F_{12}$ |
|---|---|---|---|---|---|---|
| PSO-In | 0.3285<br>(1.1587) | 21.1149<br>(0.0650) | 28.1848<br>(11.4742) | 214.2491<br>(84.8990) | 38.6029<br>(7.9234) | 3.0743e+004<br>(2.9043e+004) |
| PSO-Co | 0.0283<br>(0.0184) | 21.1271<br>(0.0557) | 71.0598<br>(22.0534) | 123.1232<br>(51.0717) | 26.6597<br>(5.1673) | 1.0415e+004<br>(1.3897e+004) |
| PSO-In-Lbest | 0.1830<br>(0.1093) | 20.9274<br>(0.0518) | 39.0149<br>(8.0007) | 149.9040<br>(39.4806) | 29.4701<br>(2.2549) | 1.6420e+004<br>(8.2755e+003) |
| SPSO<br>(PSO-Co-Lbest) | 0.0108<br>(0.0078) | **20.9092**<br>**(0.0592)** | 65.1992<br>(13.3166) | 90.4544<br>(18.4968) | 29.1374<br>(2.1661) | 4.5191e+003<br>(3.3662e+003) |
| GBBPSO | 0.0179<br>(0.0170) | 20.9631<br>(0.0481) | 60.3143<br>(15.3916) | 127.2546<br>(48.5001) | 28.2383<br>(3.4455) | 1.7318e+004<br>(6.4095e+004) |
| FIPS | 0.0147<br>(0.0101) | 20.9638<br>(0.0476) | 47.9595<br>(9.9315) | 170.4301<br>(19.0757) | 32.6119<br>(2.5941) | 3.1169e+004<br>(1.5581e+004) |
| DMS-PSO | 0.0283<br>(0.0226) | 20.9569<br>(0.0522) | 29.5427<br>(7.4630) | 77.6689<br>(11.9670) | 23.8535<br>(2.1849) | 7.4986e+003<br>(6.2259e+003) |
| CLPSO | 1.0054<br>(0.0663) | 20.9613<br>(0.0499) | **7.3197e-006**<br>**(1.2443e-005)** | 118.2419<br>(14.6277) | 23.8084<br>(2.1761) | 3.4442e+004<br>(7.6392e+003) |
| RDPSO-Gbest | 0.0175<br>(0.0140) | 20.9558<br>(0.0641) | 22.7650<br>(5.7728) | 78.6024<br>(38.9282) | 21.6689<br>(7.6173) | 6.0361e+003<br>(5.1260e+003) |
| RDPSO-Gbest-RP | 0.0130<br>(0.0123) | 20.9602<br>(0.0569) | 31.9085<br>(8.7969) | 82.5152<br>(47.7362) | **20.0701**<br>**(6.9879)** | **2.8227e+003**<br>**(3.3963e+003)** |
| RDPSO-Lbest | **0.0092**<br>**(0.0050)** | 20.9540<br>(0.0508) | 27.8237<br>(6.0386) | **49.8606**<br>**(12.9486)** | 22.1984<br>(3.0396) | 4.0616e+003<br>(2.8103e+003) |
| RDPSO-Lbest-RP | 0.0093<br>(0.0061) | 20.9613<br>(0.0543) | 36.4589<br>(7.9391) | 51.4390<br>(7.9391) | 23.0731<br>(1.8929) | 3.7315e+003<br>(1.9675e+003) |

Table 4. Mean and Standard Deviation of the Best Fitness Values after 100 runs of Each Algorithm for $F_{13}$ to $F_{18}$

| Algorithms | $F_{13}$ | $F_{14}$ | $F_{15}$ | $F_{16}$ | $F_{17}$ | $F_{18}$ |
|---|---|---|---|---|---|---|
| PSO-In | 5.2896<br>(5.5476) | 13.8002<br>(0.3444) | 316.2599<br>(111.0781) | 335.4358<br>(113.4565) | 335.4358<br>(113.4565) | 835.7976<br>(6.0356) |
| PSO-Co | 4.4108<br>(1.2793) | 12.7952<br>(0.4972) | 394.4609<br>(79.4799) | 225.5245<br>(143.6150) | 286.3587<br>(153.8241) | 833.2801<br>(3.5208) |
| PSO-In-Lbest | 5.1283<br>(1.3492) | 13.0249<br>(0.2546) | 235.7861<br>(33.0672) | 191.0510<br>(34.8487) | 236.8845<br>(32.0096) | 835.2138<br>(3.3085) |
| SPSO<br>(PSO-Co-Lbest) | 4.1371<br>（0.8434） | 12.6110<br>（0.2924） | 339.5646<br>(67.1847) | 137.7149<br>(34.8816) | 208.1448<br>(49.4636) | 834.7874<br>(2.9012) |
| GBBPSO | 4.9260 (1.3859) | 13.5393 (0.5470) | 455.5165 (108.0158) | 303.7125 (127.5004) | 371.2200 (147.7408) | 838.1823 (5.3269) |
| FIPS | 8.4372 (1.3535) | 12.7804 (0.2627) | 288.9921 (41.2366) | 245.1298 (50.0474) | 283.5242 (61.8710) | 837.6627 (2.0417) |
| DMS-PSO | 5.1709 (1.7631) | 12.6673 (0.3139) | 384.2450 (103.9382) | 135.6315 (77.8942) | 207.6865 (101.0296) | 830.4012 (2.0887) |
| CLPSO | 3.8576 (0.3906) | 13.1524 (0.1691) | **201.6172 (43.9150)** | 218.5159 (32.2153) | 272.4979 (36.2060) | 848.7486 (2.6709) |
| RDPSO-Gbest | 3.6656 (1.6923) | 12.4291 (0.4285) | 319.7043 (111.0345) | 256.3401 (142.5937) | 271.5714 (168.7019) | **827.3072 (3.6785)** |
| RDPSO-Gbest-RP | 4.1837 (2.9678) | 12.5091 (0.4033) | 332.5331 (118.6331) | 238.4040 (158.6436) | 296.8933 (191.0346) | 827.7514 (1.7776) |
| RDPSO-Lbest | 3.5717 (1.1045) | 12.4730 (0.2576) | 247.1870 (60.9730) | 87.7190 (24.6532) | **130.1013 (48.4074)** | 830.3490 (1.7263) |
| RDPSO-Lbest-RP | **3.3328 (0.7410)** | **12.3497 (0.7410)** | 230.5609 (44.6314) | **81.6018 (14.6833)** | 131.6444 (31.0220) | 834.0747 (1.7544) |

Table 5. Mean and Standard Deviation of the Best Fitness Values after 100 runs of Each Algorithm for $F_{18}$ to $F_{25}$

| Algorithms | $F_{19}$ | $F_{20}$ | $F_{21}$ | $F_{22}$ | $F_{23}$ | $F_{24}$ | $F_{25}$ |
|---|---|---|---|---|---|---|---|
| PSO-In | 832.5137 (3.9585) | 833.8981 (5.9478) | 871.1860 (5.1597) | 521.2325 (6.5959) | 873.3791 (4.0608) | 214.6238 (1.9056) | 214.1518 (1.2947) |
| PSO-Co | 834.4989 (3.7228) | 833.1274 (3.1489) | 870.7109 (124.2766) | 611.7368 (120.9721) | 839.4523 (141.4307) | 234.1746 (50.4443) | 236.4987 (3.4339) |
| PSO-In-Lbest | 836.5300 (3.4975) | 835.6854 (2.7836) | 871.4610 (2.4423) | 520.7401 (3.2509) | 873.1013 (2.2462) | 214.7839 (1.4485) | 215.4278 (1.3933) |
| SPSO (PSO-Co-Lbest) | 835.0596 (3.1710) | 834.4713 (4.6952) | 536.5353 (26.5527) | 541.4679 (13.9804) | 864.4379 (55.6264) | 219.2533 (2.3766) | 220.2320 (2.4519) |
| GBBPSO | 838.8955 (4.7992) | 838.3364 (4.4266) | 878.7394 (10.1300) | 579.0716 (93.1903) | 917.7662 (100.7194) | 224.8365 (5.2437) | 224.1366 (4.4376) |
| FIPS | 837.7127 (2.5667) | 837.4681 (2.5232) | 534.3545 (135.9025) | 757.1625 (84.2028) | 535.4757 (4.9815) | 208.8901 (18.0803) | 201.9316 (1.5447) |
| DMS-PSO | 830.1580 (2.2565) | 830.6192 (2.4581) | 840.3275 (92.5244) | **514.0093 (1.2493)** | 869.5029 (8.6424) | 211.8410 (0.5409) | 212.0612 (0.6642) |
| CLPSO | 848.9094 (3.0432) | 848.8751 (2.8792) | 806.1130 (108.8076) | 613.3743 (33.7734) | 853.3524 (36.4153) | 226.7629 (1.2682) | 231.9183 (2.2955) |
| RDPSO-Gbest | **827.2807 (2.1788)** | **827.8117 (2.4067)** | 575.1067 (163.7323) | 523.5113 (8.6283) | 632.2784 (204.7378) | **202.6466 (1.2486)** | **200.1250 (1.1607)** |
| RDPSO-Gbest-RP | 828.4142 (2.0324) | 827.9494 (1.8825) | 606.5063 (194.7164) | 522.2864 (9.3827) | 642.4793 (206.3007) | 203.4066 (1.1849) | 200.4140 (1.4218) |
| RDPSO-Lbest | 829.9966 (1.2966) | 830.2114 (1.2753) | **510.8933 (8.5183)** | 521.9427 (4.6614) | **534.1642 (2.8262e-004)** | 203.3198 (1.1920) | 201.3821 (1.3847) |
| RDPSO-Lbest-RP | 834.4578 (2.5108) | 834.0832 (1.6421) | 512.2468 (12.1296) | 528.0377 (3.4471) | 534.1644 (3.3058e-004) | 203.2279 (1.2006) | 200.8703 (1.0484) |

Table 6. T Values and P Values of the Unpaired T Test between Two Adjacent Mean Best Fitness Values in Ascending Order, with 0.05 as the Level of Significance ($F_1$ to $F_5$)

| $F_1$ | | $F_2$ | | $F_3$ | | $F_4$ | | $F_5$ | |
|---|---|---|---|---|---|---|---|---|---|
| RDPSO-Lbest-RP ~RDPSO-Gbest-RP | 17.2502<br><0.0001 | PSO-Co ~GBBPSO | 0.1047<br>0.9167 | RDPSO-Gbest-RP ~SPSO | 3.7252<br>0.0003 | RDPSO-Gbest-RP ~RDPSO-Gbest | 3.0500<br>0.0026 | RDPSO-Gbest-RP~ DMS-PSO | 2.0245<br>0.0443 |
| RDPSO-Gbest-RP ~FIPS | 4.9489<br><0.0001 | GBBPSO~ RDPSO-Gbest | 5.1396<br><0.0001 | SPSO~ RDPSO-Gbest | 3.8210<br>0.0002 | RDPSO-Gbest ~PSO-Co | 2.6441<br>0.0088 | DMS-PSO~ RDPSO-Gbest | 2.1363<br>0.0339 |
| FIPS ~SPSO | 11.9049<br><0.0001 | RDPSO-Gbest ~ RDPSO-Gbest-RP | 0.3182<br>0.7506 | RDPSO-Gbest~ RDPSO-Lbest-RP | 0.2802<br>0.7796 | PSO-Co ~DMS-PSO | 0.8517<br>0.3954 | RDPSO-Gbest~ RDPSO-Lbest | 1.3208<br>0.1881 |
| SPSO ~RDPSO-Lbest | 4.1314<br><0.0001 | RDPSO-Gbest-RP ~FIPS | 0.2544<br>0.7994 | RDPSO-Lbest-RP ~GBBPSO | 0.2864<br>0.7749 | DMS-PSO ~GBBPSO | 0.5783<br>0.5637 | RDPSO-Lbest ~ FIPS | 7.1651<br><0.0001 |
| RDPSO-Lbest ~ PSO-Co | 6.2465<br><0.0001 | FIPS~ SPSO | 10.1408<br><0.0001 | GBBPSO~ RDPSO-Lbest | 0.2352<br>0.8143 | GBBPSO ~RDPSO-Lbest | 4.1288<br><0.0001 | FIPS~ RDPSO-Lbest-RP | 4.8120<br><0.0001 |
| PSO-Co ~PSO-In | 5.7811<br><0.0001 | SPSO ~RDPSO-Lbest | 8.2914<br><0.0001 | RDPSO-Lbest ~DMS-PSO | 1.7898<br>0.0750 | RDPSO-Lbest ~PSO-In | 5.1056<br><0.0001 | RDPSO-Lbest-RP~ CLPSO | 1.2509<br>0.2124 |
| PSO-In ~RDPSO-Gbest | 26.4615<br><0.0001 | RDPSO-Lbest~ RDPSO-Lbest-RP | 10.0498<br><0.0001 | DMS-PSO ~FIPS | 3.6645<br>0.0003 | PSO-In~RDPSO-Lbest-RP | 2.6580<br>0.0085 | CLPSO ~ PSO-In | 2.2396<br>0.0262 |
| RDPSO-Gbest ~BBPSO | 2.4745<br>0.0142 | RDPSO-Lbest-RP ~DMS-PSO | 18.5569<br><0.0001 | FIPS ~PSO-Co | 4.7998<br><0.0001 | RDPSO-Lbest-RP ~FIPS | 5.4290<br><0.0001 | PSO-In ~SPSO | 4.4695<br><0.0001 |
| BBPSO ~CLPSO | 17.9072<br><0.0001 | DMS-PSO ~PSO-In | 1.9935<br>0.0476 | PSO-Co ~PSO-In-Lbest | 7.0651<br><0.0001 | FIPS ~SPSO | 6.5369<br><0.0001 | SPSO ~ PSO-Co | 4.5227<br><0.0001 |
| CLPSO ~DMS-PSO | 3.9022<br><0.0001 | PSO-In ~ PSO-In-Lbest | 8.4676<br><0.0001 | PSO-In-Lbest ~CLPSO | 5.7050<br><0.0001 | SPSO ~CLPSO | 4.3554<br><0.0001 | PSO-Co~ PSO-In-Lbest | 7.7325<br><0.0001 |
| DMS-PSO ~PSO-In-Lbest | 5.2711<br><0.0001 | PSO-In-Lbest ~CLPSO | 8.4471<br><0.0001 | CLPSO ~PSO-In | 0.3168<br>0.7517 | CLPSO ~PSO-In-Lbest | 4.5332<br><0.0001 | PSO-In-Lbest ~GBBPSO | 0.0964<br>0.9233 |

Table 7. T Values and P Values of the Unpaired T Test between Two Adjacent Mean Best Fitness Values in Ascending Order, with 0.05 as the Level of Significance ($F_6$ to $F_{10}$)

| $F_6$ | $F_7$ | $F_8$ | $F_9$ | $F_{10}$ |
|---|---|---|---|---|

| | | | | | | | | | |
|---|---|---|---|---|---|---|---|---|---|
| RDPSO-Lbest~ | 1.5181 | RDPSO-Lbest~ | 0.1268 | SPSO | 2.3137 | CLPSO | 39.4349 | RDPSO-Lbest | 1.0392 |
| RDPSO-Lbest-RP | 0.1306 | RDPSO-Lbest-RP | 0.8992 | ~PSO-In-Lbest | 0.0217 | ~ RDPSO-Gbest | <0.0001 | ~ RDPSO-Lbest-RP | 0.3000 |
| RDPSO-Lbest-RP | 2.3698 | RDPSO-Lbest-RP | 1.5148 | PSO-In-Lbest | 3.6663 | RDPSO-Gbest | 6.0554 | RDPSO-Lbest-RP | 18.2647 |
| ~RDPSO-Gbest-RP | 0.0188 | ~SPSO | 0.1314 | ~ RDPSO-Lbest | 0.0003 | ~ RDPSO-Lbest | <0.0001 | ~ DMS-PSO | <0.0001 |
| RDPSO-Gbest-RP | 1.4005 | SPSO ~ | 1.5105 | RDPSO-Lbest | 0.2201 | RDPSO-Lbest~ | 0.2785 | DMS-PSO | 0.2292 |
| ~ SPSO | 0.1629 | RDPSO-Gbest-RP | 0.1325 | ~ RDPSO-Gbest | 0.8260 | PSO-In | 0.7809 | ~ RDPSO-Gbest | 0.8189 |
| SPSO | 0.9301 | RDPSO-Gbest-RP | 1.0681 | RDPSO-Gbest | 0.1331 | PSO-In~ | 0.9921 | RDPSO-Gbest | 0.6352 |
| ~ PSO-In-Lbest | 0.3535 | ~ FIPS | 0.2868 | ~ DMS-PSO | 0.8943 | DMS-PSO | 0.3224 | ~ RDPSO-Gbest-RP | 0.5260 |
| PSO-In-Lbest | 0.0034 | FIPS | 1.6220 | DMS-PSO~ | 0.4274 | DMS-PSO~ | 2.0508 | RDPSO-Gbest-RP | 1.5508 |
| ~ PSO-Co | 0.9973 | ~ RDPSO-Gbest | 0.1064 | RDPSO-Gbest-RP | 0.6696 | RDPSO-Gbest-RP | 0.0416 | ~SPSO | 0.1225 |
| PSO-Co | 0.2903 | RDPSO-Gbest | 0.1816 | RDPSO-Gbest-RP | 0.1453 | RDPSO-Gbest-RP | 3.8401 | SPSO | 11.7835 |
| ~ RDPSO-Gbest | 0.7719 | ~GBBPSO | 0.8561 | ~ CLPSO | 0.8846 | ~ RDPSO-Lbest-RP | 0.0002 | ~ CLPSO | <0.0001 |
| RDPSO-Gbest | 1.5804 | GBBPSO | 4.1515 | CLPSO~ | 0.0000 | RDPSO-Lbest-RP | 2.2677 | CLPSO | 0.9188 |
| ~ CLPSO | 0.1156 | ~ PSO-Co | <0.0001 | RDPSO-Lbest-RP | 1.0000 | ~ PSO-In-Lbest | 0.0244 | ~ PSO-Co | 0.3593 |
| CLPSO | 1.9072 | PSO-Co | 0.0000 | RDPSO-Lbest-RP | 0.2481 | PSO-In-Lbest~ | 7.0136 | PSO-Co | 0.5866 |
| ~ FIPS | 0.0579 | ~ DMS-PSO | 1.0000 | ~ GBBPSO | 0.8043 | FIPS | <0.0001 | ~GBBPSO | 0.5582 |
| FIPS | 0.0193 | DMS-PSO | 13.8605 | GBBPSO | 0.9177 | FIPS~GBBPSO | 6.7448 | GBBPSO | 3.6217 |
| ~GBBPSO | 0.9847 | ~ PSO-In-Lbest | <0.0001 | ~ FIPS | 0.1034 | | <0.0001 | ~ PSO-In-Lbest | 0.0004 |
| GBBPSO | 0.7177 | PSO-In-Lbest | 1.2502 | FIPS | 18.7550 | GBBPSO~SPSO | 2.4001 | PSO-In-Lbest | 4.6813 |
| ~ PSO-In | 0.4738 | ~ PSO-In | 0.2127 | ~ PSO-In | <0.0001 | | 0.0173 | ~ FIPS | <0.0001 |
| PSO-In | 1.4610 | PSO-In | 5.8324 | PSO-In | 1.4252 | SPSO~ PSO-Co | 2.2749 | FIPS | 5.0358 |
| ~ DMS-PSO | 0.1456 | ~ CLPSO | <0.0001 | ~ PSO-Co | 0.1557 | | 0.0240 | ~ PSO-In | <0.0001 |

Table 8. T Values and P Values of the Unpaired T Test between Two Adjacent Mean Best Fitness Values in Ascending Order, with 0.05 as the Level of Significance ($F_{11}$ to $F_{15}$)

| $F_{11}$ | | $F_{12}$ | | $F_{13}$ | | $F_{14}$ | | $F_{15}$ | |
|---|---|---|---|---|---|---|---|---|---|
| RDPSO-Gbest-RP | 1.5467 | RDPSO-Gbest-RP | 2.3154 | RDPSO-Lbest-RP~ | 1.7962 | RDPSO-Lbest-RP~ | 0.9276 | CLPSO~ | 4.6226 |
| ~ RDPSO-Gbest | 0.1235 | ~ RDPSO-Lbest-RP | 0.0216 | RDPSO-Lbest | 0.0740 | RDPSO-Gbest | 0.3547 | RDPSO-Lbest-RP | <0.0001 |
| RDPSO-Gbest~ | 0.6456 | RDPSO-Lbest-RP | 0.9622 | RDPSO-Lbest~ | 0.4647 | RDPSO-Gbest~ | 0.8781 | RDPSO-Lbest-RP~ | 0.9407 |
| RDPSO-Lbest | 0.5193 | ~ RDPSO-Lbest | 0.3371 | RDPSO-Gbest | 0.6427 | RDPSO-Lbest | 0.3810 | PSO-In-Lbest | 0.3480 |
| RDPSO-Lbest~ | 2.4427 | RDPSO-Lbest | 1.0433 | RDPSO-Gbest~ | 1.1055 | RDPSO-Lbest~ | 0.7544 | PSO-In-Lbest~ | 1.6437 |
| RDPSO-Lbest-RP | 0.0155 | ~SPSO | 0.2981 | CLPSO | 0.2703 | RDPSO-Gbest-RP | 0.4515 | RDPSO-Lbest | 0.1018 |
| RDPSO-Lbest-RP | 2.5494 | SPSO | 2.4737 | CLPSO | 3.0071 | RDPSO-Gbest-RP~ | 2.0456 | RDPSO-Lbest | 5.6794 |
| ~ CLPSO | 0.0115 | ~ RDPSO-Gbest | 0.0142 | ~SPSO | 0.0030 | SPSO | 0.0421 | ~ FIPS | <0.0001 |
| CLPSO | 0.1463 | RDPSO-Gbest | 1.8135 | SPSO~ | 0.1510 | SPSO | 1.3124 | FIPS | 2.3014 |
| ~ DMS-PSO | 0.8839 | ~ DMS-PSO | 0.0713 | RDPSO-Gbest-RP | 0.8801 | ~ DMS-PSO | 0.1909 | ~ PSO-In | 0.0224 |
| DMS-PSO | 5.0019 | DMS-PSO | 1.9152 | RDPSO-Gbest-RP | 0.7027 | DMS-PSO | 2.7631 | PSO-In | 0.2193 |
| ~ PSO-Co | <0.0001 | ~ PSO-Co | 0.0569 | ~ PSO-Co | 0.4831 | ~ FIPS | 0.0063 | ~ RDPSO-Gbest | 0.8266 |
| PSO-Co | 2.5418 | PSO-Co | 3.7127 | PSO-Co | 2.7316 | FIPS | 0.2632 | RDPSO-Gbest~ | 0.7895 |
| ~GBBPSO | 0.0118 | ~ PSO-In-Lbest | 0.0003 | ~ GBBPSO | 0.0069 | ~ PSO-Co | 0.7927 | RDPSO-Gbest-RP | 0.4308 |
| GBBPSO | 2.2092 | PSO-In-Lbest | 0.1390 | GBBPSO ~ | 1.0459 | PSO-Co~ | 4.1121 | RDPSO-Gbest-RP | 0.5157 |
| ~SPSO | 0.0283 | ~GBBPSO | 0.8896 | PSO-In-Lbest | 0.2969 | PSO-In-Lbest | <0.0001 | ~SPSO | 0.6066 |
| SPSO | 1.0640 | GBBPSO | 2.0999 | PSO-In-Lbest~ | 0.1919 | PSO-In-Lbest~ | 4.1716 | SPSO | 3.6102 |
| ~ PSO-In-Lbest | 0.2886 | ~ FIPS | 0.0370 | DMS-PSO | 0.8480 | CLPSO | <0.0001 | ~ DMS-PSO | 0.0004 |
| PSO-In-Lbest | 9.1407 | FIPS | 0.1293 | DMS-PSO ~ | 0.2039 | CLPSO | 6.7576 | DMS-PSO | 0.7808 |
| ~ FIPS | <0.0001 | ~ PSO-In | 0.8973 | PSO-In | 0.8386 | ~GBBPSO | <0.0001 | ~ PSO-Co | 0.4359 |
| FIPS | 7.1858 | PSO-In | 1.8861 | PSO-In | 5.5121 | GBBPSO | 4.0363 | PSO-Co | 4.5528 |
| ~ PSO-In | <0.0001 | ~CLPSO | 0.0607 | ~ FIPS | <0.0001 | ~ PSO-In | <0.0001 | ~GBBPSO | <0.0001 |

Table 9. T Values and P Values of the Unpaired T Test between Two Adjacent Mean Best Fitness Values in Ascending Order ,with 0.05 as the Level of Significance ($F_{16}$ to $F_{20}$)

| $F_{16}$ | | $F_{17}$ | | $F_{18}$ | | $F_{19}$ | | $F_{20}$ | |
|---|---|---|---|---|---|---|---|---|---|
| RDPSO-Lbest-RP | 2.1318 | RDPSO-Lbest | 0.2684 | RDPSO-Gbest | 1.0873 | RDPSO-Gbest | 3.8042 | RDPSO-Gbest | 0.4507 |
| ~ RDPSO-Lbest | 0.0343 | ~ RDPSO-Lbest-RP | 0.7887 | ~ RDPSO-Gbest-RP | 0.2782 | ~ RDPSO-Gbest-RP | 0.0002 | ~ RDPSO-Gbest-RP | 0.6527 |
| RDPSO-Lbest | 5.8643 | RDPSO-Lbest-RP | 7.1952 | RDPSO-Gbest-RP~ | 10.4831 | RDPSO-Gbest-RP | 6.5639 | RDPSO-Gbest-RP | 9.9481 |
| ~ DMS-PSO | <0.0001 | ~ DMS-PSO | <0.0001 | RDPSO-Lbest | <0.0001 | ~ RDPSO-Lbest | <0.0001 | ~ RDPSO-Lbest | <0.0001 |
| DMS-PSO | 0.2441 | DMS-PSO | 0.0407 | RDPSO-Lbest | 0.1926 | RDPSO-Lbest | 0.6202 | RDPSO-Lbest~ | 1.4726 |
| ~SPSO | 0.8074 | ~SPSO | 0.9675 | ~DMS-PSO | 0.8474 | ~ DMS-PSO | 0.5359 | DMS-PSO | 0.1424 |
| SPSO | 10.8172 | SPSO | 4.8780 | DMS-PSO | 7.0324 | DMS-PSO | 5.1700 | DMS-PSO | 6.2788 |
| ~ PSO-In-Lbest | <0.0001 | ~ PSO-In-Lbest | <0.0001 | ~ PSO-Co | <0.0001 | ~ PSO-In | <0.0001 | ~ PSO-Co | <0.0001 |
| PSO-In-Lbest | 5.7872 | PSO-In-Lbest | 2.0201 | PSO-Co~ | 2.0200 | PSO-In~ | 4.1473 | PSO-Co | 1.1452 |
| ~ CLPSO | <0.0001 | ~ RDPSO-Gbest | 0.0447 | RDPSO-Lbest-RP | 0.0447 | RDPSO-Lbest-RP | <0.0001 | ~ PSO-In | 0.2535 |
| CLPSO | 0.4762 | RDPSO-Gbest | 0.0537 | RDPSO-Lbest-RP | 2.1021 | RDPSO-Lbest-RP | 0.0915 | PSO-In~ | 0.3000 |
| ~ PSO-Co | 0.6345 | ~ CLPSO | 0.9572 | ~SPSO | 0.0368 | ~ PSO-Co | 0.9272 | RDPSO-Lbest-RP | 0.7645 |
| PSO-Co~ | 0.6019 | CLPSO | 1.5381 | SPSO | 0.9690 | PSO-Co | 1.1466 | RDPSO-Lbest-RP | 0.7802 |
| RDPSO-Gbest-RP | 0.5480 | ~ FIPS | 0.1256 | ~ PSO-In-Lbest | 0.3337 | ~SPSO | 0.2529 | ~SPSO | 0.4362 |
| RDPSO-Gbest-RP | 0.4043 | FIPS | 0.1710 | PSO-In-Lbest | 0.8482 | SPSO | 3.1146 | SPSO~ | 2.2243 |
| ~ FIPS | 0.6864 | ~ PSO-Co | 0.8644 | ~ PSO-In | 0.3974 | ~ PSO-In-Lbest | 0.0021 | PSO-In-Lbest | 0.0273 |
| FIPS | 0.7418 | PSO-Co~ | 0.4295 | PSO-In | 2.9272 | PSO-In-Lbest | 2.7262 | PSO-In-Lbest | 4.7450 |
| ~ RDPSO-Gbest | 0.4591 | RDPSO-Gbest-RP | 0.6680 | ~ FIPS | 0.0038 | ~ FIPS | 0.0070 | ~ FIPS | <0.0001 |
| RDPSO-Gbest | 2.4766 | RDPSO-Gbest-RP | 1.7347 | FIPS | 0.9108 | FIPS~ | 2.1733 | FIPS~ | 1.7041 |
| ~GBBPSO | 0.0141 | ~ PSO-In | 0.0844 | ~GBBPSO | 0.3635 | GBBPSO | 0.0309 | GBBPSO | 0.0899 |
| GBBPSO | 1.8587 | PSO-In | 1.9210 | GBBPSO | 17.7317 | GBBPSO | 17.6216 | GBBPSO~ | 19.9574 |
| ~ PSO-In | 0.0645 | ~GBBPSO | 0.0562 | ~ CLPSO | <0.0001 | ~ CLPSO | <0.0001 | CLPSO | <0.0001 |

Table 10. T Values and P Values of the Unpaired T Test between Two Adjacent Mean Best Fitness Values in Ascending

Order, with 0.05 as the Level of Significance ($F_{20}$ to $F_{25}$)

| $F_{21}$ | | $F_{22}$ | | $F_{23}$ | | $F_{24}$ | | $F_{25}$ | |
|---|---|---|---|---|---|---|---|---|---|
| RDPSO-Lbest~ | 0.9132 | DMS-PSO | 19.3265 | RDPSO-Lbest | 4.5985 | RDPSO-Gbest | 3.3559 | RDPSO-Gbest | 1.5746 |
| RDPSO-Lbest-RP | 0.3623 | ~PSO-In-Lbest | <0.0001 | ~RDPSO-Lbest-RP | <0.0001 | ~ RDPSO-Lbest-RP | 0.0009 | ~ RDPSO-Gbest-RP | 0.1170 |
| RDPSO-Lbest-RP | 1.6203 | PSO-In-Lbest | 0.6696 | RDPSO-Lbest-RP | 2.6323 | RDPSO-Lbest-RP~ | 0.5432 | RDPSO-Gbest-RP~ | 2.5830 |
| ~ FIPS | 0.1068 | ~ PSO-In | 0.5039 | ~ FIPS | 0.0091 | RDPSO-Lbest | 0.5876 | RDPSO-Lbest-RP | 0.0105 |
| FIPS | 0.1575 | PSO-In~ | 0.8793 | FIPS~ | 4.7267 | RDPSO-Lbest | 0.5164 | RDPSO-Lbest-RP~ | 2.9468 |
| ~SPSO | 0.8750 | RDPSO-Lbest | 0.3803 | RDPSO-Gbest | <0.0001 | ~ RDPSO-Gbest-RP | 0.6061 | RDPSO-Lbest | 0.0036 |
| SPSO~ | 2.3254 | RDPSO-Lbest~ | 0.3281 | RDPSO-Gbest | 0.3510 | RDPSO-Gbest-RP~ | 3.0264 | RDPSO-Lbest | 2.6488 |
| RDPSO-Gbest | 0.0211 | RDPSO-Gbest-R P | 0.7432 | ~RDPSO-Gbest-RP | 0.7260 | FIPS | 0.0028 | ~ FIPS | 0.0087 |
| RDPSO-Gbest~ | 1.2342 | RDPSO-Gbest-R | 0.9609 | RDPSO-Gbest-RP~ | 7.8750 | FIPS~ | 1.6314 | FIPS~ | 60.2434 |
| RDPSO-Gbest-RP | 0.2186 | P~ RDPSO-Gbest | 0.3378 | PSO-Co | <0.0001 | DMS-PSO | 0.1044 | DMS-PSO | <0.0001 |
| RDPSO-Gbest-RP | 8.9488 | RDPSO-Gbest~ | 4.8716 | PSO-Co | 0.9518 | DMS-PSO | 14.0483 | DMS-PSO | 14.3671 |
| ~ CLPSO | <0.0001 | RDPSO-Lbest-RP | <0.0001 | ~ CLPSO | 0.3424 | ~ PSO-In | <0.0001 | ~ PSO-In | <0.0001 |
| CLPSO | 2.3955 | RDPSO-Lbest-RP | 9.3271 | CLPSO | 1.6673 | PSO-In~ | 0.6689 | PSO-In~ | 6.7088 |
| ~ DMS-PSO | 0.0175 | ~SPSO | <0.0001 | ~SPSO | 0.0970 | PSO-In-Lbest | 0.5044 | PSO-In-Lbest | <0.0001 |
| DMS-PSO | 1.9610 | SPSO~ | 3.9905 | SPSO~ | 0.8997 | PSO-In-Lbes | 16.0583 | PSO-In-Lbest | 17.0354 |
| ~ PSO-Co | 0.0513 | GBBPSO | <0.0001 | DMS-PSO | 0.3693 | ~SPSO | <0.0001 | ~SPSO | <0.0001 |
| PSO-Co | 0.0382 | GBBPSO | 2.1391 | DMS-PSO ~ | 4.0298 | SPSO~ | 9.6979 | SPSO | 7.7015 |
| ~ PSO-In | 0.9696 | ~ PSO-Co | 0.0337 | PSO-In-Lbest | 0.0001 | GBBPSO | <0.0001 | ~GBBPSO | <0.0001 |
| PSO-In | 0.4817 | PSO-Co~ | 0.1304 | PSO-In-Lbest | 0.5986 | GBBPSO~ | 3.5708 | GBBPSO | 15.5754 |
| ~ PSO-In-Lbest | 0.6305 | CLPSO | 0.8964 | ~ PSO-In | 0.5501 | CLPSO | 0.0004 | ~ CLPSO | <0.0001 |
| PSO-In-Lbest | 6.9849 | CLPSO | 15.8491 | PSO-In | 4.4034 | CLPSO~ | 1.4688 | CLPSO | 11.0892 |
| ~GBBPSO | <0.0001 | ~ FIPS | <0.0001 | ~GBBPSO | <0.0001 | PSO-Co | 0.1435 | ~ PSO-Co | <0.0001 |

Table 11. Ranking by Algorithms and Problems Obtained from "Stepdown" Multiple Comparisons

| Algorithms | $F_1$ | $F_2$ | $F_3$ | $F_4$ | $F_5$ | $F_6$ | $F_7$ | $F_8$ | $F_9$ | $F_{10}$ | $F_{11}$ | $F_{12}$ | $F_{13}$ |
|---|---|---|---|---|---|---|---|---|---|---|---|---|---|
| PSO-In | 7 | =9 | =11 | 7 | 8 | =9 | 11 | 15 | =3 | 12 | 12 | =10 | =8 |
| PSO-Co | 6 | =1 | 9 | 5 | 10 | 7 | =8 | 15 | 12 | =7 | 7 | 7 | =5 |
| PSO-In-Lbest | 12 | 11 | 10 | 12 | =11 | =3 | 10 | =1 | =7 | 10 | 10 | =8 | =8 |
| SPSO | 4 | 6 | 2 | 10 | 9 | =3 | =1 | =1 | =10 | =3 | =8 | 4 | =5 |
| GBBPSO | 9 | =1 | =3 | =3 | =11 | =9 | =4 | =3 | =10 | =7 | =8 | =8 | =8 |
| FIPS | 3 | =3 | 8 | 9 | 5 | =9 | =4 | =3 | 9 | 11 | 11 | =10 | 12 |
| DMS-PSO | 11 | =9 | =3 | =3 | =1 | 12 | =8 | =3 | =3 | =3 | =4 | =5 | =8 |
| CLPSO | 10 | 12 | =11 | 11 | =6 | =8 | 12 | =3 | 1 | =7 | =4 | =10 | 4 |
| RDPSO-Gbest | 8 | =3 | =3 | 2 | =1 | =3 | =4 | =3 | 2 | =3 | =1 | =5 | =1 |
| RDPSO-Gbest-RP | 2 | =3 | 1 | 1 | =1 | =3 | =4 | =3 | 6 | =3 | =1 | =1 | =5 |
| RDPSO-Lbest | 5 | 7 | =3 | 6 | 4 | =1 | =1 | =3 | =3 | =1 | =1 | =1 | =1 |
| RDPSO-Lbest-RP | 1 | 8 | =3 | 8 | =6 | =1 | =1 | =3 | =7 | =1 | =4 | =1 | =1 |

| Algorithms | $F_{14}$ | $F_{15}$ | $F_{16}$ | $F_{17}$ | $F_{18}$ | $F_{19}$ | $F_{20}$ | $F_{21}$ | $F_{22}$ | $F_{23}$ | $F_{24}$ | $F_{25}$ | *Ave.Rank* |
|---|---|---|---|---|---|---|---|---|---|---|---|---|---|
| PSO-In | 12 | =6 | =11 | 11 | =7 | 5 | =5 | =9 | =2 | 11 | =7 | 7 | 8.60 |
| PSO-Co | =7 | =10 | =6 | =6 | 5 | =6 | =5 | =9 | =10 | =6 | =11 | 12 | 7.68 |
| PSO-In-Lbest | 9 | =2 | 5 | 5 | =7 | 9 | 9 | =9 | =2 | 10 | =7 | 8 | 7.80 |
| SPSO | =5 | =6 | =3 | =3 | =7 | =6 | 8 | 4 | 8 | =6 | 9 | 9 | 5.60 |
| GBBPSO | 11 | 12 | =11 | 12 | =10 | 11 | =10 | 12 | 9 | 12 | 10 | 10 | 8.56 |
| FIPS | =7 | 5 | 9 | =6 | =10 | 10 | =10 | =1 | 12 | 3 | =5 | 5 | 7.20 |
| DMS-PSO | =5 | 10 | =3 | =3 | =3 | =3 | =3 | 8 | 1 | 9 | =5 | 6 | 5.28 |
| CLPSO | 10 | 1 | =6 | =6 | 12 | 12 | 12 | 7 | =10 | =6 | =11 | 11 | 8.12 |
| RDPSO-Gbest | =1 | =6 | 10 | =6 | =1 | 1 | =1 | =5 | =4 | =4 | 1 | =1 | 3.20 |
| RDPSO-Gbest-RP | =1 | =6 | =6 | =6 | =1 | 2 | =1 | =5 | =4 | =4 | =2 | =1 | 2.92 |
| RDPSO-Lbest | =1 | 4 | 2 | =1 | =3 | =3 | =3 | =1 | =4 | 1 | =2 | 4 | 2.64 |
| RDPSO-Lbest-RP | =1 | =2 | 1 | =1 | 6 | =6 | =5 | =1 | 7 | 2 | =2 | 3 | 3.28 |

For the four RDPSO variants, the linearly decreasing $\alpha$ with fixed $\beta$ was used, and the parameters for each case were set as those indentified and recommended by the previous experiments. Thus, the results for the first twelve functions by the RPSO variants are those listed in Table 1, which were yielded by the algorithms with linearly varying $\alpha$. The parameter configurations for other PSO variants were the same as those recommended by the existing publications. For the PSO-In, the inertia weight linearly decreased from 0.9 to 0.4 in the course of the run and we fixed the acceleration coefficients ($c_1$ and $c_2$) at 2.0, as in the empirical study performed by Shi and Eberhart (1999). For the PSO-Co, the constriction factor was set to be $\chi = 0.7298$, and the acceleration coefficients $c_1$=$c_2$=2.05, as recommended by Clerc and Kennedy (2002).

Eberhart and Shi also used these values of the parameters when comparing the performance of the PSO-Co with that of the PSO-In (Eberhart and Shi, 2000). For the SPSO, the ring topology was used and other parameters were set as those in the PSO-Co (Bratton and Kennedy, 2007). Parameter configurations for the GBBPSO, FIPS, DMS-PSO and CLPSO were the same as those in (Kennedy, 2003; Mendes et al., 2004; Liang and Suganthan, 2005; Liang et al., 2006), respectively.

Tables 2 and 5 record the mean and the standard deviation of the best fitness values out of 100 runs of each algorithm on each problem. To investigate if the differences in the mean best fitness values between the algorithms were significant, two statistical tests were employed for this purpose. The first one is the unpaired $t$ test between two adjacent mean best fitness values in ascending order for each problem with 0.05 as the level of significance. The $t$ values and $p$ values of the $t$ tests are provided in Tables 6 to 10. The second one is a multiple comparison procedure which was used to determine the algorithmic performance ranking for each problem in a statistical manner. The procedure employed in this work is known as the "stepdown" procedure (Day and Quinn, 1989). The algorithms that were not statistically different to each other were given the same rank; those that were not statistically different to more than one other groups of algorithms were ranked with the best-performing of these groups. For each algorithm, the resulting rank for each problem and the average rank across all the twenty five benchmark problems are shown in Table 11.

For the Shifted Sphere Function ($F_1$), the RDPSO-Lbest-RP generated better results than the other methods. The results for the Shifted Schwefel's Problem 1.2 ($F_2$) show that the PSO-Co and the GBBPSO performed better than the others, but the performance of the CLPSO seems to be inferior to those of other competitors due to its slow convergence speed. For the Shifted Rotated High Conditioned Elliptic Function ($F_3$), the RDPSO-Gbest-RP outperformed the other methods in a statistical significance manner. The SPSO was the second best performing method for this function. The RDPSO-Gbest-RP showed to be the winner among all the tested algorithms for the Shifted Schwefel's Problem 1.2 with Noise in Fitness ($F_4$), and the RDPSO-Gbest was the second best performing for this problem. $F_5$ is the Schwefel's Problem 2.6 with Global Optimum on the Bounds. For this benchmark, the RDPSO-Gbest-RP occupied the first place from the perspective of the statistical test. For benchmark $F_6$, the Shifted Rosenbrock Function, both the RDPSOs with the ring topology outperformed the other algorithms. The results for the Shifted Rotated Griewank's Function without Bounds ($F_7$) suggest that both the RDPSOs with local best model and the SPSO were able to find the solution to the function with better quality compared to the other methods. Benchmark $F_8$ is the Shifted Rotated Ackley's Function with Global Optimum on the Bounds. The SPSO and the PSO-In-Lbest yielded better results for this problem than the others. The Shifted Rastrigin's Function ($F_9$) is a separable function, which the CLPSO algorithm was good at solving and obtained remarkably better results for. It can also be observed that the RPDOS-Gbest yielded a better result than the remainders. $F_{10}$ is the Shifted Rotated Rastrigrin's Function, which appears to be a more difficult problem than $F_9$. For this benchmark, both the RDPSO-Lbest and RDPSO-Lbest-RP outperformed the other competitors in a statistically significant manner. The best result for the Shifted Rotated Weierstrass Function ($F_{11}$) was obtained by the RDPSO-Gbest-RP. The RDPSO-Gbest yielded the second best result which shows no statistical significance with that of the RDPSO-Gbest-RP. When searching the optima of Schewefel's Problem 2.13 ($F_{12}$), the RDPSO-Gbest-RP was found to rank first in algorithmic performance from a statistical point of view.

$F_{13}$ is the Shifted Expand Griewank's plus Rosenbrock's Function, for which the RDPSO-Lbest-RP, RDPSO-Lbest, and RDPSO-Gbest yielded better results than their competitors. There are no statistically significant differences in algorithmic performance between the three RDPSO variants. For the Shifted Rotated Expanded Scaffer's $F6$ Function ($F_{14}$), all the RDPSO variants showed better performance than the others in a statistically significant manner. $F_{15}$ is a hybrid composition function. For this function, the CLPSO performed the best among all the tested algorithms. The results for $F_{16}$, the rotated version of $F_{15}$, indicate that the RDPSO-Lbest-RP had the best performance in solving this problem than its competitors. $F_{17}$ is the $F_{16}$ with noise in fitness and the two version of the RDPSO with the local best model showed significant advantages over the other algorithms for this problem. As for $F_{18}$, a rotated hybrid composition function, the two versions of the RDPSO with the global best model performed the best as shown by the results. Similar conclusions can be found for $F_{19}$ (the Rotated Hybrid Composition Function with narrow basin global optimum) and $F_{20}$ (the Rotated Hybrid Composition Function with Global Optimum on the Bounds). $F_{21}$ is another rotated hybrid composition function and the RDPSO-Lbest and RDPSO-Lbest-RP shared the first place with the FIPS in algorithmic performance for this problem. For the Rotated Hybrid Composition Function with High Condition Number Matrix ($F_{22}$), the performance of DMS-PSO was superior to those of the other methods. The result obtained by the RDPSO-Lbest for $F_{23}$, the Non-Continuous Rotated Hybrid Composition Function, is statistically the best among all the algorithms. For $F_{24}$ (another rotated hybrid composition function) and $F_{25}$ (the version of $F_{24}$ without bounds), the RDPSO-Gbest-RP and RDPSO-Gbest showed ~~the~~ better performance than the others.

The average ranks listed in Table 11 reveal that the RDPSO-Lbest had the best overall performance among all the tested algorithms. For ten of the benchmark functions, the algorithm had the first performance ranks, but unsatisfactory performance for $F_2$ due to its slow convergence speed resulted from the local best model. The second best-performing was the RPDSO-Gbest-RP. Across the whole suite of benchmark functions, it had fairly stable performance, with the worst rank being 6 for $F_9$, $F_{16,}$ $F_{17}$ and $F_{18}$, respectively. The RDPSO-Gbest had the third best overall performance. Compared to the RDPSO-Gbest-RP, the RDPSO-Gbest performed somewhat unstable, with the resulting ranks for $F_1$ and $F_{16}$ being only 8 and 10, respectively. However, like the RDPSO-Gbest-RP, the RDPSO-Gbest had the first ranks for nine benchmark functions. The fourth best performing was the RDPSO-Lbest-RP, which did not show satisfactory performance on $F_2$ and $F_4$. Nevertheless, it had a significant advantage over the DMS-PSO, the next best performing one. Between random velocity components determined by the *mbest* position and the random selected *pbest* position, the two versions of the RDPSO with the *mbest* position obtained the total average rank of 2.92, while the two with the randomly selected *pbest* position had the total average rank 3.10. Also, what can be found from the total average ranks is that the RDPSOs were able to work slightly better by using the local best model (with the total average rank 3.06) than the global best model (with the total average ranks 3.16) for the CEC2005 benchmark suite. Besides, the total average rank over all the versions of the RDPSO is 3.11, which implies that the RDPSOs with the linearly varying $\alpha$ and fixed $\beta$ had a satisfactory overall performance. Therefore, it is recommended that the linearly varying $\alpha$ method with fixed $\beta$ should be employed when the RDPSO is used for real applications. More specifically, for the RDPSO-Gbest, RDPSO-Lbest and RDPSO-Lbest-RP, the initial value of $\alpha$ can be selected from the interval [0.8, 1.0] and the final value can be selected from the interval [0.2, 0.4] depending on the problem to be solved. For the RDPSO-Gbest-RP, the initial value of $\alpha$ can be selected from the interval [0.5, 0.7] and its final value from [0.1, 0.3]. The value of $\beta$ can be selected from the interval [1.45, 1.5] for all he RDPSO variants.

Except the RDPSO algorithms, the best-performing algorithm was the DMS-PSO, which yielded the best results for $F_{22}$. It showed good performance for most of the multimodal functions, but did not work well on unimodal ones. The next best algorithm was the SPSO, i.e. PSO-Co-Lbest. For $F_7$ and $F_8$, it yielded the best results among all the competitors. The FIPS, which also employs the ring topology, was the next best performing algorithm and generated the best results for $F_{21}$. It is conclusive, from the total average ranks in Table 11, that incorporating the ring topology into the PSO-In and the PSO-Co could enhance the overall performance of the two PSO variants on the tested benchmark functions. What should be noticed is that the CLPSO is very effective in solving separable functions such as $F_9$, but not in the rotated functions and unimodal ones due to its slower convergence speed, as has been indicated in the related publication (Liang, et al., 2006). The GBBPSO, an important probabilistic PSO variant, had good performance for unimodal functions. It is impressive that it generated the best result for $F_2$, implying that it has a relative faster convergence speed than its competitors on this type of functions.

## 5. Conclusion

In this paper, inspired by the free electron model in metal conductors in an external electric field, the RDPSO algorithm was proposed as a novel variant of PSO. The velocity updated equation of the particle in the RDPSO algorithm is a superposition of two parts, i.e., a random component and a drift component, which reflect the global and the local search of the particle, respectively. These two components were designed to simulate the thermal motion as well as the drift motion of the electron, which lead the electron to a location with minimum potential energy in a metal conductor under an external electric field. Thus, the search process of the RDPSO algorithm is analogous to the process of minimizing the electron's potential energy.

A comprehensive analysis of the RDPSO algorithm and its variants was made in order to better understand the mechanism behind the algorithm. Firstly, the stochastic dynamical behavior of a single particle in the RDPSO was analyzed theoretically. We derived the sufficient and necessary condition as well as a sufficient condition for the particle's current position to be probabilistically bounded. Secondly, the search behavior of the RDPSO algorithm itself was investigated by analyzing the interaction between the particles, and it was found that the RDPSO may have a good balance between the global and the local search, due to the designed random component of the particle's velocity. Subsequently, some variants of the RDPSO algorithm were proposed by combining different random velocity components with different neighborhood topologies.

Empirical studies on the RDPSO algorithm were conducted on all the twenty five benchmark functions of the well-known CEC2005 benchmark suite. Two methods for controlling the algorithmic parameters were employed, and each RDPSO variant, with each control method, was first tested on three of the benchmark functions in order to identify the parameter values that can generate the best performance. Then, the RDPSO variants with the identified parameter values were compared on the first twelve benchmark functions. Finally, the RDPSO variants with the linearly varying thermal coefficient and the fixed drift coefficient were further compared with other forms of PSO on all the twenty five functions. The experimental results show that the RDPSO algorithm is comparable with, or even better, than its competitors in finding the optimal solutions of the tested benchmark functions.

## Appendix

**Theorem A1:** If there is only drift motion for the particle, i.e. $V_{i,n+1}^j = VD_{i,n+1}^j$ a sufficient condition for $X_{i,n+1}^j$ to converge to $p_{i,n}^j$ is $0 < \beta < 2$.

***Proof***: From equation (14) and (16), we can find that

$$X_{i,n+1}^j - p_{i,n+1}^j = (1-\beta)(X_{i,n}^j - p_{i,n}^j) + p_{i,n}^j - p_{i,n+1}^j, \tag{A1}$$

When the RDPSO algorithm is running, the personal best positions of all the particles converge to the same point. Consequently, $\{p_{i,n}^j\}$ is a convergent Cauchy sequence such that $\lim_{n\to\infty} | p_{i,n}^j - p_{i,n+1}^j | = 0$. Since $0 < \beta < 2$, $1 - |1-\beta| > 0$. Thus, it holds that

$$\lim_{n\to\infty} [| p_{i,n}^j - p_{i,n+1}^j | / (1 - |1-\beta|)] = 0,$$

which means that for any $\varepsilon > 0$, there exists an integer $K > 0$ such that whenever $n \ge K$,

$$| p_{i,n}^j - p_{i,n+1}^j | < \varepsilon \cdot (1 - |1-\beta|). \tag{A2}$$

Therefore, from inequality (A2), we have

$$| X_{i,n+1}^j - p_{i,n+1}^j | - \varepsilon \le |1-\beta| (| X_{i,n}^j - p_{i,n}^j | - \varepsilon). \tag{A3}$$

This implies that for any $n > K$,

$$| X_{i,n}^j - p_{i,n}^j | \le \varepsilon + \left( \prod_{k=K}^{n-1} |1-\beta| \right) | X_{i,K}^j - p_{i,K}^j |. \tag{A4}$$

Since $0 \le |1-\beta| < 1$, $\lim_{n\to\infty} \prod_{k=K}^{n-1} |1-\beta| = \lim_{n\to\infty} |1-\beta|^{n-K} = 0$. Hence

$$\limsup_{n\to\infty} | X_{i,n}^j - p_{i,n}^j | \le \varepsilon. \tag{A5}$$

As $\varepsilon$ is arbitrary and $| X_{i,n}^j - p_{i,n}^j | \ge 0$, therefore

$$\lim_{n\to\infty} | X_{i,n}^j - p_{i,n}^j | = 0. \tag{A6}$$

This completes the proof of the theorem. ■

**Theorem A2:** The necessary and sufficient condition for the position sequence of the particle $\{X_n\}$ to be probabilistically bounded is that $\rho_n = \prod_{i=1}^n \lambda_i$ does not diverge, namely, $\rho_n$ is probabilistic bounded (i.e. $P\left\{\sup_{n\ge1} \rho_n < \infty\right\} = 1$).

**Proof**: From equation (17) and (18), the update equation of the particle's position is given by

$$X_{n+1} = \alpha(X_n - C)\varphi_{n+1} - \beta(X_n - p) + X_n, \tag{A7}$$

from which we immediately have

$$\begin{aligned} X_{n+1}-C &= \alpha(X_n - C)\varphi_{n+1} - \beta(X_n - p) + X_n - C \\ &= [\alpha\varphi_{n+1} + (1-\beta)](X_n - C) + \beta(p-C) = \lambda_{n+1}(X_n - C) + \beta(p-C) \end{aligned}. \tag{A8}$$

Since $\lambda_{n+1}$ is a continuous random variable, $P\{\lambda_{n+1}=1\}=0$. Considering that $\beta(p-C)$ is probabilistically bounded, we have that $r=\dfrac{\beta(p-C)}{1-\lambda_{n+1}}$ is also a probabilistic bounded random variable. From (A8), we can obtain

$$X_{n+1}-C-r = \lambda_{n+1}(X_n - C - r), \tag{A9}$$

From which we can recursively derive the following formula

$$X_n = (X_0 - C - r)\prod_{i=1}^{n}\lambda_i + C + r. \tag{A10}$$

Since $X_0 - C - r$ is probabilistically bounded, $X_n$ is probabilistic bounded if and only if $\rho_n = \prod_{i=1}^{n}\lambda_i$ is probabilistically bounded. This completes the proof of the theorem. ■

**Theorem A3:** Let $\Delta = E(\xi_n) = \dfrac{1}{\sqrt{2\pi}\alpha}\displaystyle\int_{-\infty}^{+\infty}\ln|x|\, e^{-\frac{[x-(1-\beta)]^2}{2\alpha^2}}dx$, where $\xi_n = \ln|\lambda_n|$ and $\lambda_n \sim N(1-\beta, \alpha^2)$. (1) The necessary and sufficient condition for $\rho_n = \prod_{i=1}^{n}\lambda_i$ to converge to zero with probability 1 is $\Delta < 0$. (2) The necessary and sufficient condition for $\rho_n$ to be probabilistically bounded, i.e. $P\{\sup \rho_n < \infty\} = 1$, is $\Delta \le 0$.

***Proof:*** By Kolmogorov's strong law of large numbers, it holds that

$$P\left\{\lim_{n\to\infty}\frac{1}{n}\sum_{i=1}^{n}\xi_i = E(\xi_1) = \Delta\right\} = 1, \tag{A11}$$

which is equivalent to the proposition that $\forall m \in Z^+, \exists K_1 \in Z^+$ such that whenever $k \ge K_1$,

$$\Delta - \frac{1}{m} < \frac{1}{k}\sum_{i=1}^{k}\ln|\lambda_i| < \Delta + \frac{1}{m}. \tag{A12}$$

(1) *Proof of the necessity.* If $P\{\lim_{n\to\infty}\rho_n = 0\} = 1$, we have that $P\{\lim_{n\to\infty}\ln|\rho_n| = \lim_{n\to\infty}\sum_{i=1}^{n}\ln|\lambda_i| = -\infty\} = 1$, that is, $\forall m \in Z^+, \exists K_2 \in Z^+$, such that whenever $k \ge K_2$ $\sum_{i=1}^{k}\ln|\lambda_i| < -m$ and thus

$$\frac{1}{k}\sum_{i=1}^{k}\ln|\lambda_i| < -\frac{m}{k}. \tag{A13}$$

Therefore, $\forall m \in Z^+$, there exists $K = \max(K_1, K_2)$ such that whenever $k \ge K$, both inequalities (A12) and (A13) holds, from which we have, $\Delta - 1/m < -m/k$, namely, $\Delta < 1/m - m/k$. Let $k \to \infty$, and considering the artibrariness of $1/m$, we obtain $\Delta < 0$.

*Proof of the sufficiency*. If $\Delta < 0$, from (A11) we have $P\{\lim_{n\to\infty}(1/n)\sum_{i=1}^{n}\xi_i < 0\} = 1$, which implies that $\exists\delta > 0, \exists K \in Z^+$, such that whenever $k \ge K$, $(1/k)\sum_{i=1}^{k}\ln|\lambda_i| < -\delta$, that is

$$\sum_{i=1}^{k}\ln|\lambda_i| < -k\delta. \tag{A14}$$

Due to the arbitrariness of $\delta$, we find that $\lim_{n\to\infty}\sum_{i=}^{k}\ln|\lambda_i| = -\infty$, which means that $P\{\lim_{n\to\infty}\rho_n = \lim_{n\to\infty}\prod_{i=1}^{n}\lambda_i = 0\} = 1$. This completes the proof of the sufficiency.

(2) From (A11), we have the following equivalent propositions:

$$\Delta = 0 \Leftrightarrow P\left\{\lim_{n\to\infty}\frac{1}{n}\sum_{i=1}^{n}\xi_i = 0\right\} = 1$$

$$\Leftrightarrow \forall\varepsilon > 0, \exists K \in Z^+, \text{ such that whenever } k \ge K,\ P\left\{\left|\frac{1}{k}\sum_{i=1}^{k}\xi_i\right| < \varepsilon\right\} = 1$$

$$\Leftrightarrow \forall\varepsilon > 0, \exists K \in Z^+, \text{ such that whenever } k \ge K,\ P\{-k\varepsilon < \sum_{i=1}^{k}\xi_i < k\varepsilon\} = 1$$

$$\Leftrightarrow P\{-\infty < \lim_{n\to\infty}\ln\rho_n < \infty\} \Leftrightarrow P\{0 < \lim_{n\to\infty}\rho_n < \infty\}. \tag{A15}$$

Thus, considering the case for $\Delta < 0$ in (1) and the case for $\Delta = 0$, we find that the first proposition in (1) holds.

Similarly,

$$\Delta > 0 \Leftrightarrow P\left\{\lim_{n\to\infty}\frac{1}{n}\sum_{i=1}^{n}\xi_i > 0\right\} = 1$$

$$\Leftrightarrow \exists\delta > 0,\ \exists K \in Z^+, \text{ such that whenever } k \ge K, P\left\{\frac{1}{k}\sum_{i=1}^{k}\zeta_i > \delta\right\} = 1, \text{ i.e. } P\left\{\sum_{i=1}^{k}\zeta_i > k\delta\right\} = 1$$

$$\Leftrightarrow P\left\{\lim_{n\to\infty}\sum_{i=1}^{k}\xi_i = \infty\right\} = 1 \Leftrightarrow P\{\lim_{n\to\infty}\ln\rho_n = +\infty\} = 1 \tag{A16}$$

Thus the second proposition in (2) holds.

This completes the proof of the second part of the theorem. ■

**Theorem A4:** A sufficient condition for $\rho_n = \prod_{i=1}^{n}\lambda_i$ to be probabilistically bounded is that $0 < \alpha < 1$ and $0 < \beta < 2$.

***Proof:*** Since $\{\lambda_n\}$ is a sequence of independent identically distributed (i.i.d.) random variables with each $\lambda_n$ subject to the same normal distribution, i.e., $\lambda_n \sim N(1-\beta, \alpha^2)$, the expectation and the variance of $\rho_n = \prod_{i=1}^{n}\lambda_i$ can be given by

$$E[\rho_n] = E[\prod_{i=1}^{n} \lambda_i] = [E[\lambda_n]]^n = (1-\beta)^n \,, \tag{A17}$$

and

$$Var[\rho_n] = E[(\rho_n)^2] - E[\rho_n]^2 = E[\prod_{i=1}^{n} \lambda_i^2]^n - E[\rho_n]^2 = [\alpha^2 + (1-\beta)^2]^n - (1-\beta)^{2n} \,. \tag{A18}$$

A sufficient condition for $\rho_n$ to converge is $E[\rho_n] \to 0$ and $Var[\rho_n] \to 0$ (i.e., mean square convergence of $\rho_n$), which implies that $0 < \beta < 2$ and $0 < \alpha < 1$. This completes the proof of the theorem.